\documentclass{article}

\usepackage{arxiv}
\usepackage[utf8]{inputenc}
\usepackage[T1]{fontenc}
\usepackage{hyperref}
\usepackage{url}
\usepackage{booktabs}
\usepackage{amsfonts}
\usepackage{nicefrac}
\usepackage{microtype}
\usepackage{graphicx}
\usepackage[numbers,sort&compress]{natbib}
\usepackage{doi}
\usepackage{amsmath}
\usepackage{multirow}
\usepackage{subcaption}
\usepackage{censor}

\newcommand{\etal}{\textit{et al.}}

\title{A Heterogeneous Two-Stream Framework for Video Action Recognition\\
with Comparative Fusion Analysis}

\author{
  Md. Afzalur Rahaman \\
  Department of Computer Science \& Engineering \\
  Hamdard University Bangladesh \\
  Munshiganj, Bangladesh \\
  \texttt{afzalurrahaman@hamdarduniversity.edu.bd}
  \and
  \textbf{Tahmid Rahman} \\
  Department of Computer Science \& Engineering \\
  Hamdard University Bangladesh \\
  Munshiganj, Bangladesh \\
  \texttt{tahmid@hamdarduniversity.edu.bd}
}

\renewcommand{\shorttitle}{Heterogeneous Two-Stream Framework for Action Recognition}

\hypersetup{
  pdftitle={A Heterogeneous Two-Stream Framework for Video Action Recognition with Comparative Fusion Analysis},
  pdfsubject={Computer Vision, Action Recognition},
  pdfauthor={Md. Afzalur Rahaman, Tahmid Rahman},
  pdfkeywords={Vision Transformer, MobileNetV2, Optical Flow, Action Recognition, Multimodal Fusion, Two-Stream, Deep Learning},
}

\begin{document}
\maketitle

\begin{abstract}
Most two-stream action recognition networks apply the same convolutional backbone to both RGB and optical flow streams, ignoring the fact that the two modalities have fundamentally different structural properties. Optical flow captures fine-grained motion patterns, while RGB frames carry rich appearance and scene context — treating them identically discards this distinction. We propose \emph{DualStreamHybrid}, a heterogeneous two-stream architecture that assigns each stream a backbone suited to its input: a pretrained ViT-Tiny/16 for RGB frames, and a MobileNetV2 trained from scratch on a 20-channel stacked optical flow representation. A learned projection layer maps the two differently-sized feature vectors to a common dimensionality before fusion, enabling the two streams to interact without forcing architectural symmetry. We design five fusion strategies within a unified framework — late fusion, concatenation, cross-attention, weighted fusion, and gated fusion — and evaluate them on UCF11 (1,600 videos, 11 classes) and UCF50 (6,681 videos, 50 classes) to study how fusion behaviour scales with dataset size. On UCF11, cross-attention achieves 98.12\% test accuracy, outperforming the RGB-only ViT-Tiny baseline of 95.94\%, which suggests that explicit inter-modal attention is particularly effective on smaller, less complex datasets. On UCF50, weighted fusion reaches 96.86\% and proves the most consistent strategy across both benchmarks. The learned stream weights reveal an interesting pattern: UCF11 sees near-equal modality contribution (RGB: 0.507, flow: 0.493), while UCF50 favours the RGB stream slightly more (RGB: 0.554, flow: 0.446) — arguably reflecting the larger and more visually diverse action space. Taken together, these results suggest that even a lightweight motion stream meaningfully complements a strong appearance encoder, and that the optimal fusion strategy depends on dataset scale.

\end{abstract}

\keywords{Vision Transformer \and MobileNetV2 \and Optical Flow \and Action Recognition
  \and Multimodal Fusion \and Two-Stream Architecture}

\section{Introduction}

Video-based action recognition is one of the core problems in computer vision, with applications spanning intelligent surveillance~\cite{poppe2010survey}, healthcare monitoring~\cite{aggarwal2011human}, sports analysis~\cite{baladaniya2025artificial} and human--computer interaction. What makes it genuinely difficult is that understanding an action demands both spatial and temporal information simultaneously — neither alone is sufficient. Spatial information captures scene appearance: who is present, what objects exist, what pose the person holds. Temporal information captures motion. A model relying solely on appearance can easily confuse two actions that look identical in a still frame but differ entirely in how they unfold over time. Conversely, a motion-only model discards the rich contextual detail that individual frames carry.

The two-stream paradigm, introduced by Simonyan and Zisserman~\cite{simonyan2014two}, was designed specifically to address this by running two networks in parallel: one for RGB frames, another for dense optical flow. Combining their predictions substantially outperformed either stream alone, and the core idea has since been extended in many directions~\cite{wang2016temporal,feichtenhofer2016convolutional,carreira2017quo}.Vision Transformers (ViTs)~\cite{dosovitskiy2020image} brought a different set of strengths to the table. Their self-attention mechanism captures long-range spatial relationships that convolutions inherently struggle with, and this translated into impressive results on image classification. Video transformer models — TimeSformer~\cite{bertasius2021space}, ViViT~\cite{arnab2021vivit}, and Video Swin Transformer~\cite{liu2022video} — extended these ideas to video and achieved strong performance on large benchmarks. But they generally operate on RGB sequences only. None incorporates optical flow in a two-stream manner, which leaves an open question about whether combining transformer-based appearance modelling with explicit motion signals could push performance further.

One aspect the two-stream literature has largely overlooked is the choice of backbone architecture for each stream. Almost all existing methods use the same backbone for both streams~\cite{simonyan2014two,wang2016temporal,feichtenhofer2016convolutional}, changing only the number of input channels — a design decision that, on reflection, is harder to justify than it might seem.
The two modalities process fundamentally different information. RGB frames carry globally distributed appearance cues — body pose, scene context, object presence — and ViTs handle this well precisely because self-attention can model relationships between distant image regions directly. Optical flow fields, by contrast, encode local pixel-level motion that varies smoothly across neighbouring regions. That is exactly the structure convolutional filters are built for. Applying the same backbone to both streams ignores this distinction entirely.
We argue that a heterogeneous design — one backbone per modality, chosen to match the structure of its input — is a natural and underexplored alternative. Whether it actually works better is an empirical question, and one this work sets out to answer.
A second underexplored question is how different fusion strategies compare in a controlled setting — and whether the best strategy shifts depending on dataset scale. Most two-stream papers default to late fusion, averaging the softmax scores from both streams, without seriously evaluating the alternatives.

In this paper, we try to address both of these questions. We propose
\emph{DualStreamHybrid}, a heterogeneous two-stream architecture that uses ViT-Tiny/16 for the RGB stream and MobileNetV2 for the optical flow stream. We add a learned projection layer to handle the dimensionality mismatch between the two encoders, and then implement five fusion strategies that we evaluate in a controlled way---keeping
everything except the fusion mechanism the same across experiments. We run the full comparison on both UCF11 and UCF50 to see how things change with dataset scale. An overview of the pipeline is shown in Figure~\ref{fig:overview}, and the detailed
architecture is shown in Figure~\ref{fig:architecture}.

The main contributions of this paper are:

\begin{itemize}
  \item We propose a heterogeneous two-stream architecture that assigns each stream a
  backbone suited to its input modality---ViT-Tiny/16 for RGB appearance and
  MobileNetV2 for optical flow---departing from the common practice of using the same
  architecture for both streams. A learned projection layer is used to align the
  feature dimensions before fusion, so neither backbone needs to be modified.

  \item We implement and compare five fusion strategies---late fusion, concatenation, cross-attention, weighted fusion, and gated fusion---in a controlled setting where everything except the fusion mechanism is kept identical, making the comparison fair and interpretable.

  \item We evaluate the framework on two datasets of different scales (UCF11 and UCF50), and show that the best fusion strategy is not the same for both: cross-attention works best on the smaller dataset while weighted fusion is more reliable on the larger one. This is a useful practical finding for anyone choosing a fusion strategy.

  \item The learned stream weights from weighted fusion give a direct, interpretable picture of how much each modality contributes, and we show that this balance shifts with dataset scale in a way that makes intuitive sense.
\end{itemize}

The rest of the paper is organised as follows. Section~\ref{sec:related} reviews related work on convolutional action recognition, two-stream methods, video transformers, optical
flow, and multimodal fusion. Section~\ref{sec:method} describes the proposed architecture, preprocessing pipeline, fusion strategies, and training setup. Section~\ref{sec:results} presents and discusses the experimental results, including a comparison with prior work.
Section~\ref{sec:conclusion} concludes the paper.

\section{Related Work}
\label{sec:related}

\subsection{Convolutional Networks for Action Recognition}

Deep convolutional neural networks have been central to video action recognition since Karpathy~\etal~\cite{karpathy2014large} demonstrated that CNN features could be applied to video through temporal pooling. Their work exposed a key limitation: spatial features alone are insufficient, and temporal information is essential for reliable recognition. Tran~\etal~\cite{tran2015learning} tackled this directly with C3D, which applies 3D convolutions jointly over space and time in short clips. C3D transferred well across datasets---a property that made it influential well beyond the benchmark it was introduced on. The arrival of ResNet~\cite{he2016deep} then made very deep architectures trainable through residual connections, and ResNet-based backbones quickly became the standard in two-stream pipelines. SlowFast networks~\cite{feichtenhofer2019slowfast} took a different approach altogether. Rather than treating all frames equally, they run two branches at different temporal resolutions---one to capture semantic content, another to capture fast motion---achieving strong results on Kinetics without relying on precomputed optical flow.

Handcrafted convolutional and descriptor-based methods similarly dominated early action recognition. Liu~\etal~\cite{liu2009recognizing} combined HOG and HOF descriptors with an SVM classifier, reporting 71.2\% on UCF11---an early result that established a reference point for the benchmark. Reddy and Shah~\cite{reddy2013recognizing} introduced UCF50 alongside an action bank feature baseline of 57.9\%. Yang~\etal~\cite{yang2018learning} later achieved 85.3\% on UCF11 through a block-term LSTM applied to convolutional
features, reflecting the broader shift toward combining deep spatial representations with sequential temporal modelling.

\subsection{Two-Stream Architectures}

Simonyan and Zisserman~\cite{simonyan2014two} established the two-stream paradigm by training one CNN on RGB frames and another on dense optical flow, then averaging their softmax predictions---confirming that appearance and motion carry genuinely complementary information. Wang~\etal~\cite{wang2016temporal} extended this with
temporal segment networks (TSN), which sample segments uniformly across the full
clip rather than short windows. TSN also introduced the flow normalisation protocol we adopt in this work. Feichtenhofer~\etal~\cite{feichtenhofer2016convolutional}
shifted the question from what to fuse to where, finding that merging at the last convolutional layer outperforms simple prediction averaging. What unites all three
of these works, however, is a shared assumption: both streams use the same backbone architecture, differing only in the number of input channels.

This homogeneous assumption also characterises prior two-stream methods. Gammulle~\etal~\cite{gammulle2017two} paired a two-stream VGG-16 architecture with an LSTM, achieving 94.6\% on UCF11. Mahmoud and Abdelhakim~\cite{kumar2023attention} reported 98.9\% on UCF11 and 96.04\% on UCF50 using a BiLSTM with attention, though their method operates on RGB alone without any motion stream. Our work directly challenges the homogeneous assumption by assigning each stream a backbone matched to the structural properties of its modality---ViT-Tiny for RGB appearance and MobileNetV2 for optical flow.

\subsection{Vision Transformers for Video Understanding}

Dosovitskiy~\etal~\cite{dosovitskiy2020image} showed that a transformer operating on image patches---the Vision Transformer (ViT)---can match or beat CNNs on image
classification when pretrained on sufficient data. The key advantage is self-attention: unlike convolutions, every patch can attend to every other patch
directly, making ViT naturally suited to capturing global spatial structure. Extending this to video required addressing the combinatorial explosion of spatiotemporal attention. TimeSformer~\cite{bertasius2021space} tackled this by separating spatial and temporal attention, substantially reducing computational cost. ViViT~\cite{arnab2021vivit} explored several factorisation strategies for spatiotemporal attention, finding that large-scale pretraining is critical to performance. MViT~\cite{fan2021multiscale} introduced pooled attention that progressively reduces sequence length while increasing channel depth---enabling efficient multi-scale modelling without the full quadratic cost. Video Swin Transformer~\cite{liu2022video} adapted the shifted-window design of Swin Transformer~\cite{liu2021swin} to 3D video, achieving competitive results with a more structured attention pattern.

The applicability of transformer-based encoders to smaller-scale benchmarks such as UCF50 was demonstrated by Hussain~\etal~\cite{hussain2022vision}, who coupled ViT-Base/16 with an LSTM to achieve 96.1\%---showing that transformer appearance encoders are effective even outside large-scale pretraining regimes. All of these models, however, operate exclusively on RGB sequences and do not treat optical flow as a complementary stream. Our work takes a different position: we use ViT-Tiny as a lightweight appearance encoder within a two-stream framework, pairing it with a convolutional motion encoder designed specifically for optical flow.

\subsection{Optical Flow for Action Recognition}

Reliable motion representation has always depended on the quality of optical flow estimation. Simonyan and Zisserman~\cite{simonyan2014two} demonstrated its value
for action recognition, and classical estimators like Farneback~\cite{farneback2003two} and TV-L1~\cite{zach2007tvl1} have been widely used since---with TV-L1 generally delivering better accuracy at higher computational cost. Learning-based estimators have pushed this further. RAFT~\cite{teed2020raft} uses recurrent all-pairs feature correlation to achieve substantially better accuracy on standard benchmarks, while PWC Net~\cite{sun2018pwc} takes a more efficient route through pyramid warping. Both represent a meaningful step beyond
classical methods in terms of estimation quality. One issue directly relevant to our work is how video compression affects flow quality. The MPEG and AVI files in
UCF11 and UCF50 introduce compression artefacts that degrade the inputs to classical estimators like Farneback---and this partly explains the performance ceiling observed for the optical flow stream in our experiments. It is a practical
consideration that the literature does not always acknowledge, but one that has material consequences when working with real-world compressed video collections.

\subsection{Multimodal Fusion}

How to best combine information from multiple modalities is an active area of research. Gadzicki~\etal~\cite{gadzicki2020early} compared early and late fusion for multimodal CNNs and found that the best strategy depends on how correlated the modalities are. Attention-based fusion has become popular: Almari~\etal~\cite{Alamri_2019_CVPR} used cross-modal attention for audio-visual learning, and Ilse~\etal~\cite{ilse2018attention} showed that attention-based gating can selectively weight informative inputs. In the action recognition context, Feichtenhofer~\etal~\cite{feichtenhofer2016convolutional} compared fusion at different network depths, focusing on where to fuse rather than how. As far as we know, no prior work has systematically compared five different fusion strategies within a heterogeneous ViT--CNN two-stream framework across two datasets of different scales, which is what we do here.

\section{Proposed Method}
\label{sec:method}

\subsection{Architecture Overview}

\begin{figure}[ht]
  \centering
  \includegraphics[width=0.7\linewidth]{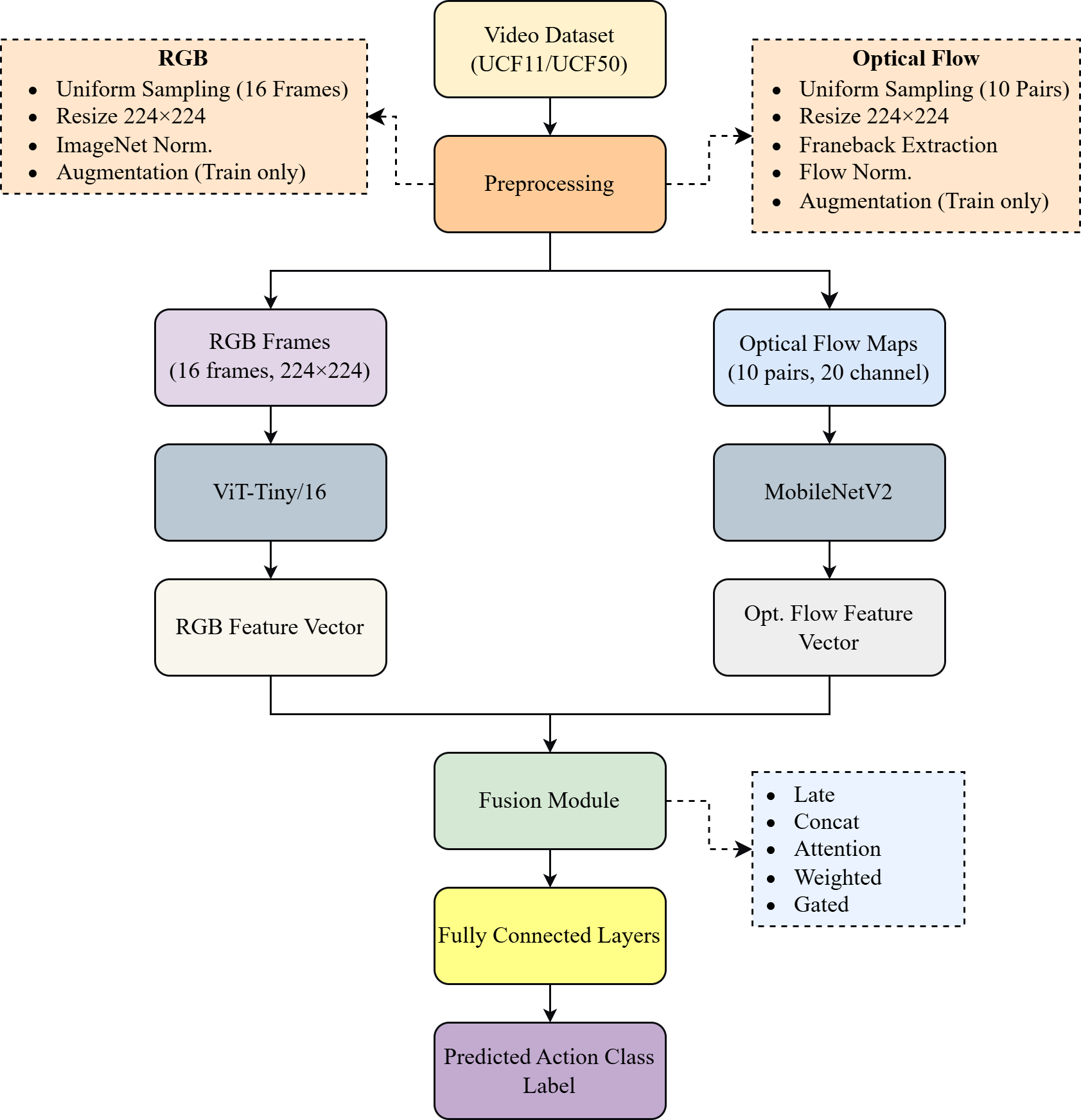}
  \caption{Overview of the proposed two-stream methodology. Video input is
    split into two preprocessing branches---RGB frames and optical flow maps---each
    fed into a dedicated encoder. The projected features are then combined by one
    of five fusion strategies before the final classification.}
  \label{fig:overview}
\end{figure}

\begin{figure}[ht]
  \centering
  \includegraphics[width=\linewidth]{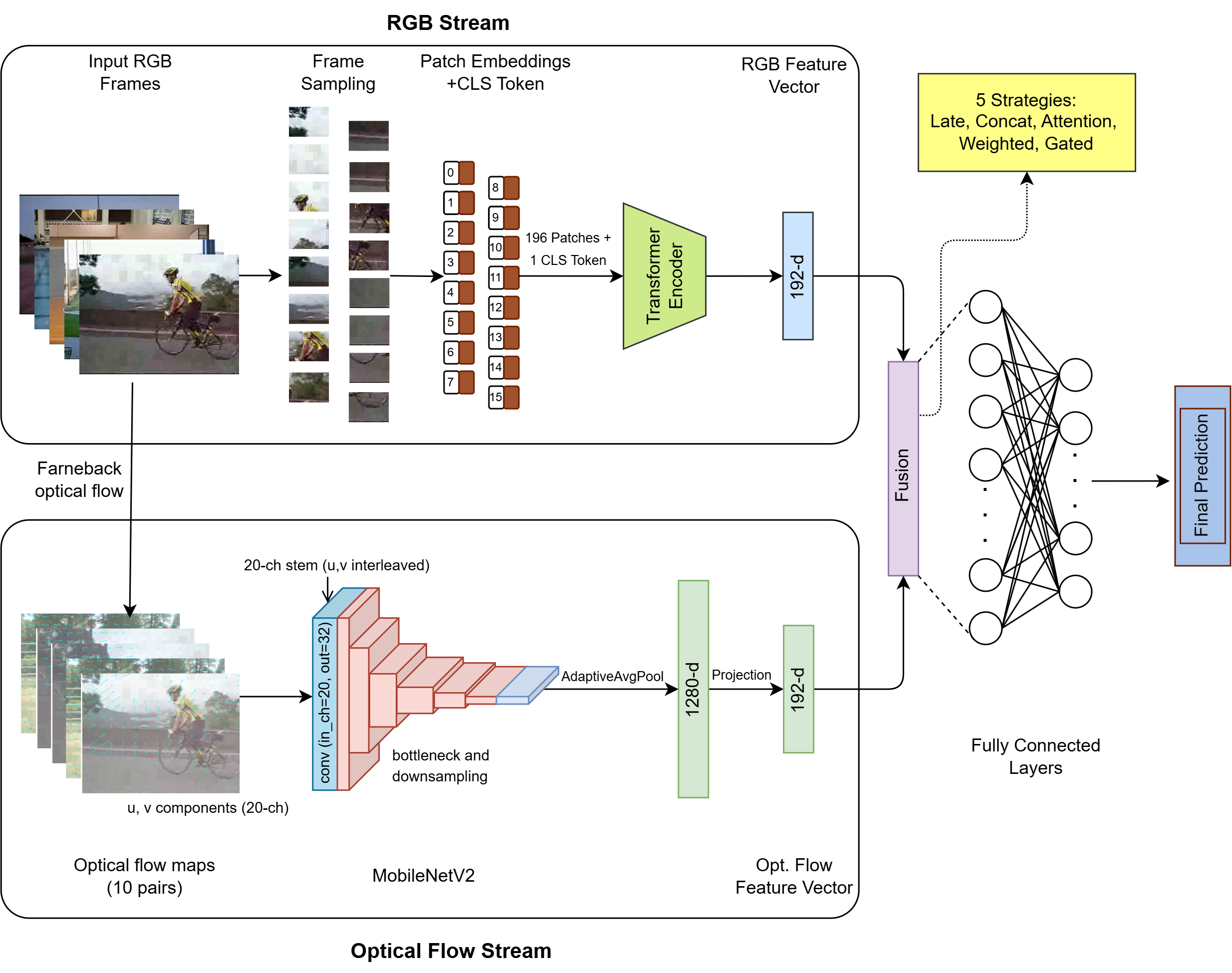}
  \caption{Detailed architecture of \emph{DualStreamHybrid}. The RGB stream
    (top) processes a single 224$\times$224 frame through ViT-Tiny/16, producing
    196 patch embeddings plus a CLS token. The flow stream (bottom) processes a
    20-channel u,v-interleaved stack through a modified MobileNetV2 (with a
    20-channel input stem, trained from scratch), followed by AdaptiveAvgPool to
    produce a 1280-d feature vector. A projection head (Linear$\to$ReLU$\to$Dropout)
    maps this to 192-d before fusion.}
  \label{fig:architecture}
\end{figure}

The proposed \emph{DualStreamHybrid} framework has three main stages, as shown in Figures~\ref{fig:overview} and~\ref{fig:architecture}. In the first stage, video
frames are uniformly sampled and optical flow is pre-computed and cached. In the second stage, a representative RGB frame is processed by the ViT-Tiny/16 encoder to get a
192-dimensional appearance feature, while the stacked flow maps are processed by MobileNetV2 to get a 1280-dimensional motion feature; a projection layer then maps both to the same 192-dimensional space. In the third stage, one of five fusion strategies combines the two aligned feature vectors to produce the final class prediction. The design keeps everything except the fusion module identical across
experiments, so the comparison between strategies is as controlled as possible.

\subsection{Preprocessing}

\subsubsection{Frame Sampling}

Given a video of $T$ frames, we uniformly sample $N = 16$ frames and resize each to $224 \times 224$ pixels. For the RGB stream, we use the middle frame $f_{\lfloor N/2 \rfloor}$ as a representative input. While this discards temporal information within the RGB stream, the optical flow stream handles temporal dynamics, and the single-frame design keeps the RGB stream simple and easy to ablate. In
practice, we found that ViT-Tiny with ImageNet pretraining provides a strong enough appearance representation from a single frame.

\subsubsection{Optical Flow Extraction}

For the motion stream, we compute dense optical flow between $K = 10$ consecutive frame pairs sampled uniformly from the 16 selected frames, using the Farneback algorithm~\cite{farneback2003two}. Farneback estimates flow via polynomial expansion of local pixel neighbourhoods, and it is computationally efficient without needing a
pretrained model. We follow the same parameter settings as TSN~\cite{wang2016temporal}: pyramid scale 0.5, five levels, window size 11, five iterations per level, polynomial
degree 5, and Gaussian smoothing $\sigma = 1.1$.

Each flow field gives horizontal ($u$) and vertical ($v$) displacement maps. Following the TSN canonical normalisation~\cite{wang2016temporal}, displacements are clipped to $\pm 20$ pixels and scaled to $[0, 255]$:
\begin{equation}
O_{\text{norm}} = \frac{\operatorname{clip}(O,\,-20,\,+20) + 20}{40} \times 255
\label{eq:flownorm}
\end{equation}
The $u$ and $v$ maps from all 10 frame pairs are interleaved into a 20-channel stack $[u_0, v_0, u_1, v_1, \ldots, u_9, v_9]$, and all channels are zero-centred by subtracting 127.5. Figure~\ref{fig:optflow} illustrates this extraction pipeline on example video clips.

\begin{figure}[ht]
  \centering
  \includegraphics[width=\linewidth]{opt_flow_viz.png}
  \caption{Optical flow extraction pipeline. Each row shows a frame pair: the raw frames at times $t$ and $t{+}1$, the HSV-encoded flow map (hue = direction, brightness = magnitude), the horizontal $u$ and vertical $v$ components after canonical normalisation, and the quiver overlay showing motion vectors. The direction key at the bottom maps hue to flow direction.}
  \label{fig:optflow}
\end{figure}

All flow stacks are pre-extracted once before training begins and saved as \texttt{.npy} files to avoid recomputing flow at every training step. A fallback that computes flow on the fly is included for any video not found in the cache.

\subsection{RGB Appearance Stream}

The RGB stream uses ViT-Tiny/16~\cite{dosovitskiy2020image} pretrained on ImageNet-21k~\cite{deng2009imagenet}. The input frame is divided into $P = 196$ non-overlapping $16 \times 16$ patches, each linearly projected to a 192-dimensional embedding. A learnable class token is prepended and learnable positional embeddings are added. The sequence is then processed by $L = 12$ transformer encoder blocks,
each with multi-head self-attention ($h = 3$ heads), layer normalisation, and a two-layer MLP with GELU activation, all with residual connections. The final class token $\mathbf{h}_\text{RGB} \in \mathbb{R}^{192}$ is the appearance feature vector.

We chose ViT over a CNN for the RGB stream because single-frame action recognition often requires reasoning about spatially distant elements — a player's position relative to a goal, the orientation of arms during a swing. Convolutions are inherently local; they cannot directly relate distant regions without stacking many layers. Self-attention has no such constraint, making ViT a more natural fit for this kind of holistic spatial reasoning. The ImageNet-21k pretraining provides a strong initialisation as well, and its effect is visible in the RGB-only baseline results reported in Section~\ref{sec:results}.

\subsection{Optical Flow Motion Stream}

The motion stream uses MobileNetV2~\cite{sandler2018mobilenetv2}, trained from
scratch on the 20-channel flow stack. MobileNetV2 is built on inverted residual blocks with linear bottlenecks, which give good representation quality at a relatively low
parameter count~\cite{simonyan2014vgg}. We replace the first convolutional layer with a 20-channel version to accept the flow input; everything else stays the same. After
the backbone, global average pooling gives a 1280-dimensional motion feature vector
$\mathbf{f}_\text{flow} \in \mathbb{R}^{1280}$.

We chose a CNN rather than a second ViT for the motion stream for two reasons. First, optical flow fields have strong local spatial coherence---neighbouring pixels on the
same moving surface tend to have nearly the same displacement---and convolutions are a natural fit for this kind of local structure. Second, there is no large-scale optical flow pretraining corpus that would let us initialise a ViT effectively; training a ViT from scratch on the small amounts of flow data in UCF11 or UCF50 would very likely overfit. MobileNetV2's smaller parameter count makes it more practical for this from-scratch training scenario.

\subsection{Projection Layer}

ViT-Tiny produces 192-dimensional features while MobileNetV2 produces 1280-dimensional features. To fuse them, we project the motion features down to the same space:
\begin{equation}
\mathbf{h}_\text{flow} = \operatorname{Dropout}_{0.1}
\!\left(\operatorname{ReLU}(\mathbf{W}_p \mathbf{f}_\text{flow} + \mathbf{b}_p)\right)
\in \mathbb{R}^{192}
\label{eq:proj}
\end{equation}
where $\mathbf{W}_p \in \mathbb{R}^{192 \times 1280}$ and $\mathbf{b}_p \in \mathbb{R}^{192}$ are learned parameters. The ReLU activation lets the projection learn a non-linear mapping rather than just a linear reduction, and the 10\% dropout provides some regularisation. This module is trained end-to-end with the rest of the network. After projection, both streams produce 192-dimensional vectors and share the
same interface for all fusion strategies.

\subsection{Fusion Strategies}

All five fusion strategies take $\mathbf{h}_\text{RGB}$ and $\mathbf{h}_\text{flow}$ (both in $\mathbb{R}^{192}$) and produce class logits $\mathbf{y} \in \mathbb{R}^C$,
where $C$ is the number of action classes. A shared linear classifier is used for all strategies except late fusion, which has separate heads per stream.

\subsubsection{Late Fusion}

Each stream has its own linear classifier, and the final prediction is the average of the two softmax distributions:
\begin{equation}
\mathbf{y}_\text{late} = \frac{1}{2}\Bigl(
  \operatorname{softmax}(\mathbf{W}_\text{RGB}\,\mathbf{h}_\text{RGB} + \mathbf{b}_\text{RGB})
+ \operatorname{softmax}(\mathbf{W}_\text{flow}\,\mathbf{h}_\text{flow} + \mathbf{b}_\text{flow})
\Bigr)
\label{eq:late}
\end{equation}
This allows each stream to develop its own decision boundary independently, without any cross-stream interaction. It is the simplest and most parameter-efficient of the five strategies, and it serves as a strong baseline.

\subsubsection{Concatenation Fusion}

The two 192-dimensional vectors are concatenated to form a 384-dimensional vector, which is then passed through a small two-layer MLP:
\begin{equation}
\mathbf{y}_\text{concat} = \mathbf{W}_2\,\operatorname{ReLU}
\!\left(\mathbf{W}_1\,[\mathbf{h}_\text{RGB};\,\mathbf{h}_\text{flow}] + \mathbf{b}_1\right)
+ \mathbf{b}_2
\label{eq:concat}
\end{equation}
where $\mathbf{W}_1 \in \mathbb{R}^{192 \times 384}$ maps back to 192 dimensions and $\mathbf{W}_2 \in \mathbb{R}^{C \times 192}$ is the classifier. Dropout ($p = 0.3$) is
applied after the ReLU. Concatenation allows the MLP to learn cross-modal interactions, but the additional parameters can be hard to train when data is limited.

\subsubsection{Cross-Attention Fusion}

Here, the RGB features act as the query and the projected flow features act as keys and values in a multi-head attention operation with $h_\text{attn} = 4$ heads. The
result is added back to the RGB features through a residual connection and layer normalisation:
\begin{equation}
\mathbf{h}_\text{att} = \operatorname{MHA}(\mathbf{h}_\text{RGB},\,
\mathbf{h}_\text{flow},\,\mathbf{h}_\text{flow}), \qquad
\mathbf{h}_\text{cross} = \operatorname{LN}(\mathbf{h}_\text{RGB} + \mathbf{h}_\text{att})
\label{eq:attn}
\end{equation}
and $\mathbf{y}_\text{cross} = \mathbf{W}_c\,\mathbf{h}_\text{cross} + \mathbf{b}_c$.
The residual connection means the original appearance features are preserved and not overwritten. The decision to use RGB as the query is motivated by the idea that global
appearance context---as captured by ViT---is well-positioned to decide which motion cues are relevant for a given scene.

\subsubsection{Weighted Fusion}

Two learnable scalar weights determine how much each stream contributes. They are normalised with softmax to form a convex combination:
\begin{equation}
(\alpha_\text{RGB},\,\alpha_\text{flow}) = \operatorname{softmax}([w_1,\,w_2]),
\qquad
\mathbf{h}_\text{fused} = \alpha_\text{RGB}\,\mathbf{h}_\text{RGB}
                         + \alpha_\text{flow}\,\mathbf{h}_\text{flow}
\label{eq:weighted}
\end{equation}
The fused vector is passed through a linear classifier. This strategy adds only two parameters beyond the classifier head, which makes it very parameter-efficient and
well-regularised even with limited training data. An additional benefit is that the learned $\alpha$ values directly tell us how much the model relies on each stream.

\subsubsection{Gated Fusion}

Gated fusion extends the idea of weighted fusion to a per-dimension level. A gating network processes the concatenated features through a sigmoid to produce a 192-dimensional
gate vector:
\begin{equation}
\mathbf{g} = \sigma(\mathbf{W}_g\,[\mathbf{h}_\text{RGB};\,\mathbf{h}_\text{flow}]
             + \mathbf{b}_g), \qquad
\mathbf{h}_\text{gated} = \mathbf{g} \odot \mathbf{h}_\text{RGB}
                         + (\mathbf{1} - \mathbf{g}) \odot \mathbf{h}_\text{flow}
\label{eq:gated}
\end{equation}
where $\mathbf{W}_g \in \mathbb{R}^{192 \times 384}$, $\mathbf{b}_g \in \mathbb{R}^{192}$,
and $\odot$ is element-wise multiplication. Each dimension of the fused vector is a weighted mix of the corresponding RGB and flow dimensions, giving fine-grained control
over which features come from which stream. However, gated fusion has the most additional parameters of the five strategies, which can be a disadvantage when training data is small.

\subsection{Datasets and Evaluation Protocol}

\begin{figure}[ht]
  \centering
  \includegraphics[width=\linewidth]{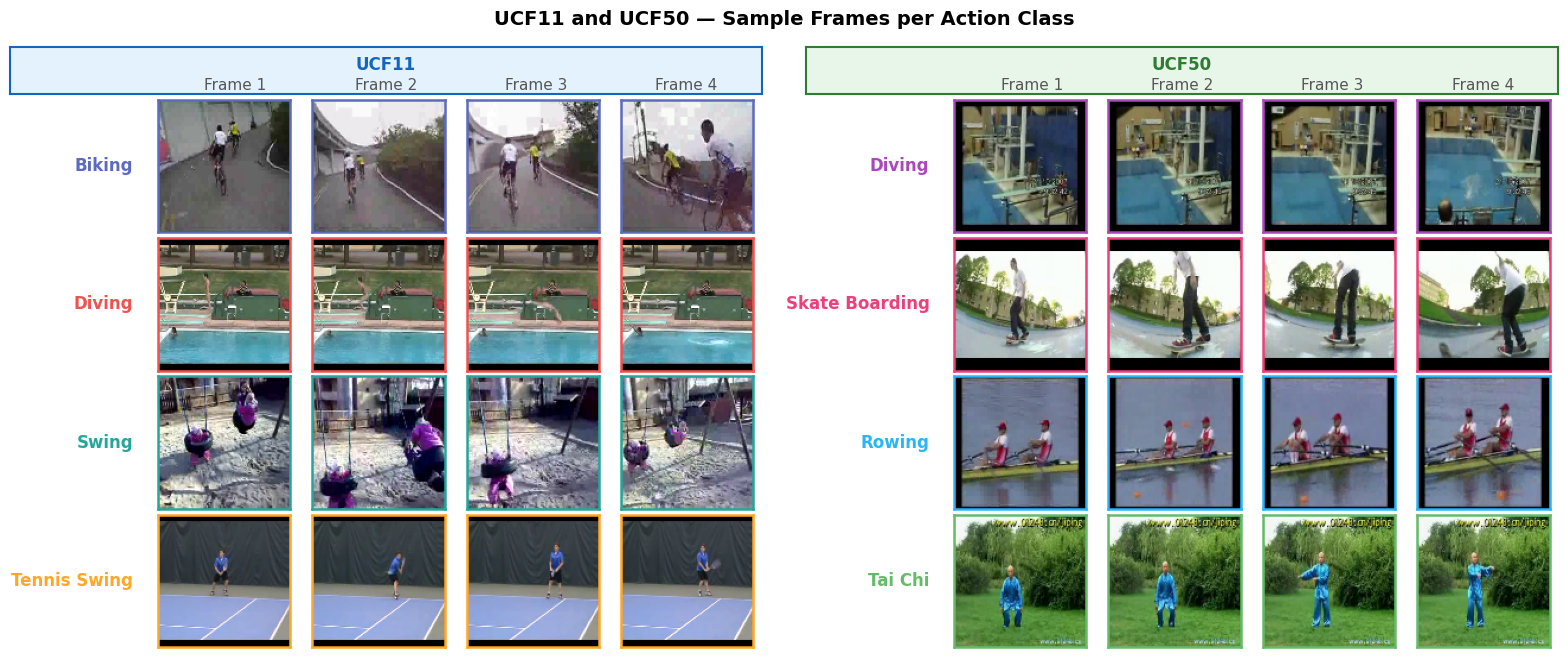}
  \caption{Sample frames from UCF11 (left) and UCF50 (right), with four uniformly sampled frames shown per action class. UCF50 shows considerably more visual variation — in camera viewpoint, background, and subject appearance — reflecting the greater diversity of its action space.}
  \label{fig:dataset}
\end{figure}

\textbf{UCF11}~\cite{liu2009recognizing}, also known as the YouTube Action dataset, contains 1,600 video clips across 11 action categories: basketball, biking, diving, golf swing, horse riding, soccer juggling, swing, tennis swing, trampoline jumping, volleyball spiking, and walking. All videos are in MPEG format. One clip in the basketball category lacked sufficient frames for optical flow extraction and was excluded, leaving 1,599 usable clips.
\textbf{UCF50}~\cite{soomro2012ucf101} extends UCF11 to 50 action categories and 6,681 videos in AVI format, covering a broader range of sports and daily activities. The scale increase is substantial — greater within-class variation makes classification considerably harder than on UCF11.
Together, the two datasets let us examine a question we consider central to this work: does the best fusion strategy change when moving from a small 11-class problem to a larger, more complex 50-class one? Figure~\ref{fig:dataset} shows sample frames from both.

We split both datasets into training (70\%), validation (10\%), and test (20\%) subsets using stratified random sampling with a fixed seed (42), ensuring proportional class representation across all splits. The validation set serves exclusively for early stopping and model selection — the test set is held out entirely and evaluated only once for final reporting.

\subsection{Training Protocol}

\subsubsection{Data Augmentation}

During training, RGB inputs are augmented with random horizontal flipping (probability 0.5) and random affine transformations (rotations up to $\pm 10^\circ$, translations up to 10\% of image size). ImageNet statistics are used for normalisation ($\boldsymbol{\mu} = [0.485, 0.456, 0.406]$, $\boldsymbol{\sigma} = [0.229, 0.224, 0.225]$). These augmentations are \emph{not} applied to optical flow inputs, because geometrically transforming a flow field would corrupt the displacement vectors it encodes. Only zero-centring is applied to flow inputs at both training and test time.

\subsubsection{Early Stopping}

We monitor validation accuracy after each epoch and apply early stopping if no improvement exceeds $10^{-4}$ over $P$ consecutive epochs---with $P = 5$ for UCF11 and $P = 4$ for UCF50. This serves two purposes. It prevents overfitting,  particularly for motion-only models trained from scratch on small datasets, and it keeps training time within the bounds of a single Kaggle GPU session across all eight experiments.

\subsubsection{Optimisation}

We use AdamW~\cite{loshchilov2017decoupled} with a cosine annealing learning rate schedule throughout. Models that include the pretrained ViT-Tiny backbone  --- the RGB-only baseline and all dual-stream models --- are trained with 
learning rate $\eta = 1 \times 10^{-4}$, weight decay 
$\lambda = 1 \times 10^{-4}$, and a maximum of 10 epochs on UCF11 or 7 epochs on UCF50. The flow-only model is different. Trained entirely from scratch, it needs more time to converge, so we use a higher learning rate $\eta = 5 \times 10^{-4}$ and allow up to 50 epochs on UCF11 or 30 epochs on UCF50. All models use a batch size of 8 and categorical cross-entropy loss. The checkpoint with the best validation accuracy is retained for test evaluation.

\section{Experiments}
\label{sec:results}

\subsection{Single-Stream Baselines}

Before examining the fusion results, we first evaluate each stream in isolation to understand what each contributes on its own. These results are reported in Tables~\ref{tab:ucf11} and~\ref{tab:ucf50}.

\textbf{RGB-only.} The pretrained ViT-Tiny achieves 95.94\% on UCF11 and 95.81\% on UCF50. The near-identical accuracy across both datasets is arguably the more interesting finding here --- UCF50 has 4.5 times more action classes, 
yet the gap is negligible. This likely reflects how well ImageNet-pretrained features generalise: the model recognises actions from a single frame reliably, even with 50 classes in play. Training converges quickly on both datasets and early stopping never triggers, suggesting the model does not overfit within the allowed epoch budget.

\textbf{Flow-only.} MobileNetV2 trained from scratch on the 20-channel flow stack achieves 49.69\% on UCF11 and 59.99\% on UCF50. The improvement on the larger dataset is expected --- more training examples help the model learn discriminative flow patterns without any pretraining to fall back on. UCF11 tells a different story. Early stopping triggers at epoch 21 (out of 50) because validation accuracy plateaued at 51.25\% while training accuracy kept climbing --- a clear sign of overfitting. On UCF50, early stopping triggers at epoch 29 (out of 30), near the end of the training budget.

\begin{table}[ht]
\centering
\caption{Classification results on \textbf{UCF11} (1,599 videos, 11 classes). Dual-stream models use ViT-Tiny/16 (RGB, pretrained ImageNet-21k) and MobileNetV2 (flow, trained from scratch) with a learned projection layer. Early stopping patience = 5. Best results per column in \textbf{bold}.}
\label{tab:ucf11}
\begin{tabular}{llccc}
\toprule
\textbf{Experiment} & \textbf{Type} & \textbf{Val Acc (\%)} & \textbf{Test Acc (\%)} & \textbf{Macro F1 (\%)} \\
\midrule
RGB-only         & Single & 96.88 & 95.94 & 95.87 \\
Flow-only        & Single & 51.25 & 49.69 & 49.31 \\
\midrule
Late fusion      & Dual   & 98.75 & 96.25 & 96.18 \\
Concat fusion    & Dual   & 98.12 & 97.81 & 97.74 \\
Attention fusion & Dual   & \textbf{99.38} & \textbf{98.12} & \textbf{98.07} \\
Weighted fusion  & Dual   & 97.50 & 96.88 & 96.80 \\
Gated fusion     & Dual   & 98.75 & 97.81 & 97.75 \\
\bottomrule
\end{tabular}
\end{table}

\begin{table}[ht]
\centering
\caption{Classification results on \textbf{UCF50} (6,681 videos, 50 classes). Same
  architecture as UCF11. Early stopping patience = 4. Maximum epochs: 7 (pretrained
  models), 30 (flow-only). Best results per column in \textbf{bold}.}
\label{tab:ucf50}
\begin{tabular}{llccc}
\toprule
\textbf{Experiment} & \textbf{Type} & \textbf{Val Acc (\%)} & \textbf{Test Acc (\%)} & \textbf{Macro F1 (\%)} \\
\midrule
RGB-only         & Single & 94.16 & 95.81 & 95.74 \\
Flow-only        & Single & 60.48 & 59.99 & 60.02 \\
\midrule
Late fusion      & Dual   & 94.61 & 95.21 & 95.11 \\
Concat fusion    & Dual   & 94.61 & 95.51 & 95.49 \\
Attention fusion & Dual   & \textbf{94.76} & 96.48 & 96.43 \\
Weighted fusion  & Dual   & 94.61 & \textbf{96.86} & \textbf{96.83} \\
Gated fusion     & Dual   & \textbf{94.76} & 95.36 & 95.25 \\
\bottomrule
\end{tabular}
\end{table}

\subsection{Dual-Stream Fusion Results}

\textbf{UCF11.} Every fusion strategy improves over the RGB-only baseline on UCF11, confirming that MobileNetV2 motion features carry useful complementary information even though the flow encoder is not particularly strong in isolation. 
Late fusion --- which simply averages the two output distributions --- already reaches 96.25\%, showing that even without feature-level interaction, two independently trained classifiers can produce meaningfully complementary 
predictions. Concatenation and gated fusion both reach 97.81\%.

Cross-attention achieves the highest test accuracy of 98.12\%, alongside the best validation accuracy of 99.38\%. This result makes intuitive sense. Using the ViT appearance feature as a query to attend over flow features is a natural 
fit for this architecture --- the globally-contextualised ViT representation is well-positioned to select which motion cues matter for a given scene. Weighted fusion reaches 96.88\% with learned weights $\alpha_\text{RGB} = 0.507$ and 
$\alpha_\text{flow} = 0.493$, nearly equal. This near-balanced weighting is direct evidence that both modalities contribute roughly the same amount on UCF11.

\begin{figure}[ht]
  \centering
  \begin{subfigure}[b]{0.48\linewidth}
    \centering
    \includegraphics[width=\linewidth]{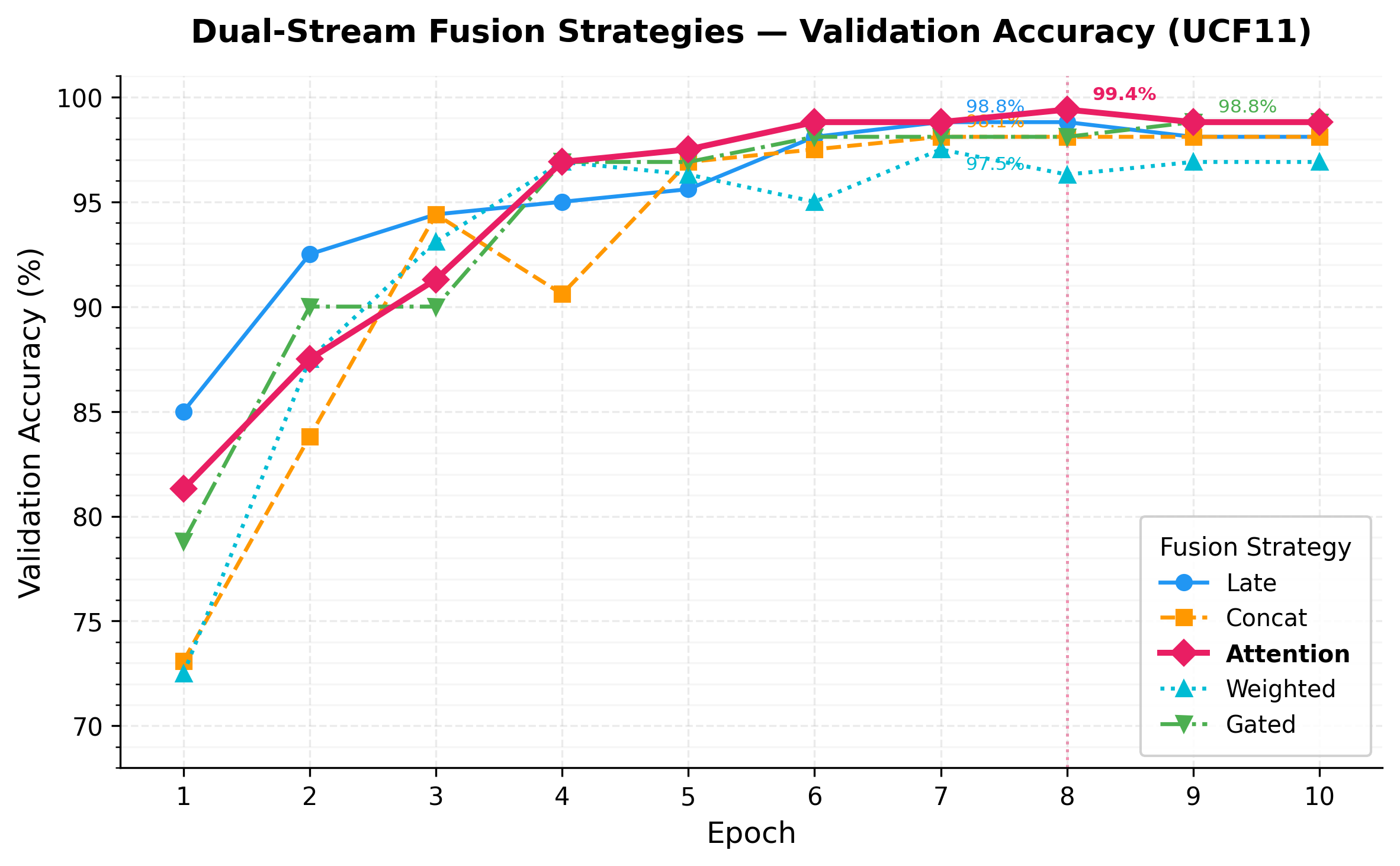}
    \caption{UCF11}
    \label{fig:curves_ucf11}
  \end{subfigure}
  \hfill
  \begin{subfigure}[b]{0.48\linewidth}
    \centering
    \includegraphics[width=\linewidth]{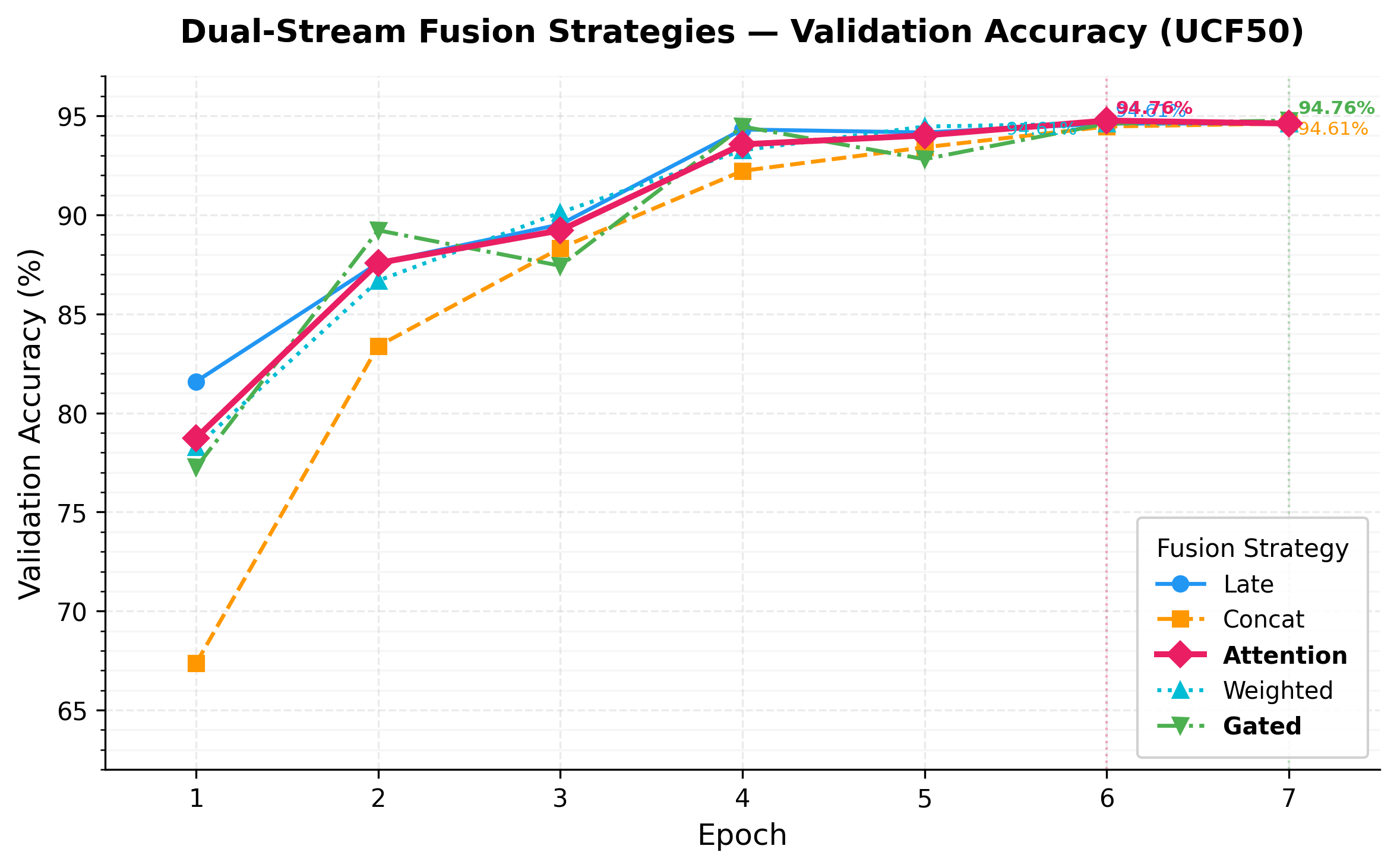}
    \caption{UCF50}
    \label{fig:curves_ucf50}
  \end{subfigure}

  \caption{Validation accuracy curves for dual-stream fusion strategies. (Left) UCF11: Cross-attention achieves the highest peak accuracy (99.38\%) at epoch~8. (Right) UCF50: Performance converges closely, with attention and weighted fusion yielding the strongest results.}
  
  \label{fig:curves_combined}
\end{figure}

\begin{figure}[ht]
  \centering
  \begin{subfigure}[b]{0.48\linewidth}
    \centering
    \includegraphics[width=\linewidth]{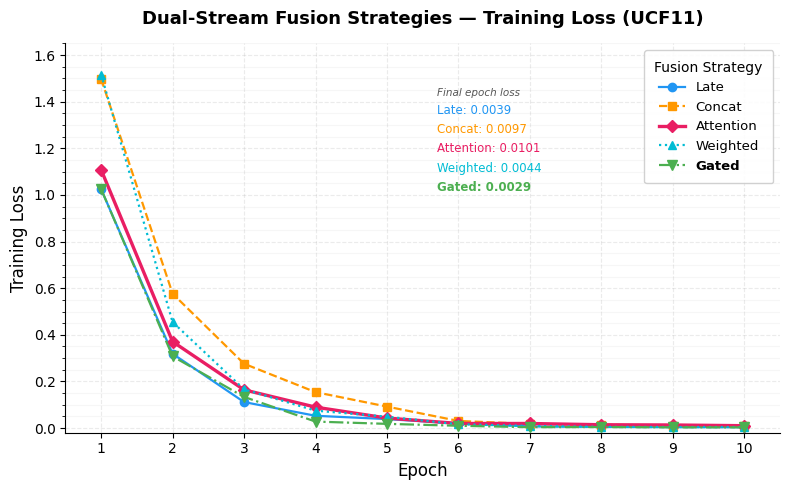}
    \caption{UCF11}
    \label{fig:loss_ucf11}
  \end{subfigure}
  \hfill
  \begin{subfigure}[b]{0.48\linewidth}
    \centering
    \includegraphics[width=\linewidth]{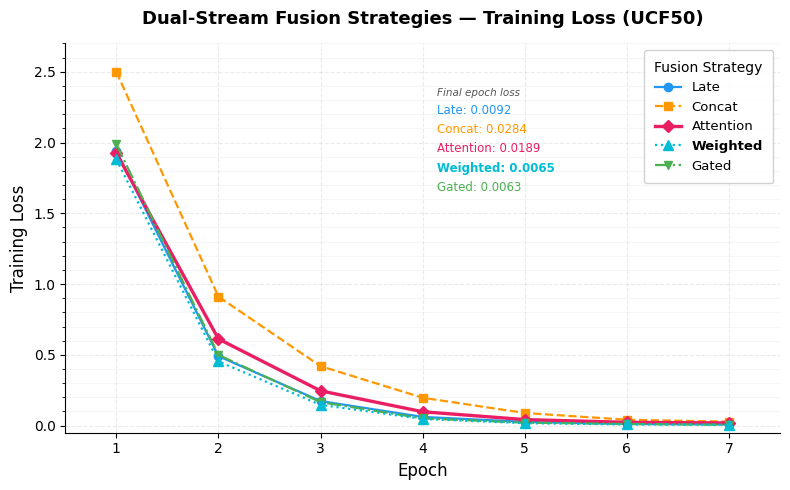}
    \caption{UCF50}
    \label{fig:loss_ucf50}
  \end{subfigure}

  \caption{Training loss curves for dual-stream fusion strategies. (Left) UCF11: All methods converge rapidly, with gated fusion achieving the lowest final loss (0.0029), followed by weighted fusion. (Right) UCF50: Weighted and gated fusion obtain the lowest final losses, 
  indicating improved optimization stability compared to other strategies.}
  
  \label{fig:loss_combined}
\end{figure}

\textbf{UCF50.} The picture changes on UCF50. Late fusion (95.21\%), concatenation (95.51\%), and gated fusion (95.36\%) all fall below the RGB-only baseline of 95.81\% --- a reversal that is worth examining carefully. These strategies carry more fusion parameters that need to be learned, and the 7-epoch budget on a harder 50-class problem is arguably not enough for them to converge fully. Cross-attention (96.48\%) and weighted fusion (96.86\%) are the only two strategies that improve over the baseline.
Weighted fusion's learned weights shift to $\alpha_\text{RGB} = 0.554$ and 
$\alpha_\text{flow} = 0.446$, giving the RGB stream a slightly larger role than on UCF11. This is unsurprising. Many of the 50 action classes involve recognizable objects or equipment --- musical instruments, bicycles, weights --- their appearance alone can identify action reliably. When the action space is this visually diverse, the RGB stream naturally carries more discriminative weight.
Figure~\ref{fig:curves_ucf50} shows validation curves on UCF50. The strategies converge tightly in the 94--95\% range, with weighted fusion and cross-attention reaching slightly higher final values.
Figures~\ref{fig:curves_combined} and~\ref{fig:loss_combined} present the validation accuracy and training loss trajectories for all five dual-stream fusion strategies. Across both datasets, convergence occurs quickly, typically within a small number of epochs, which underscores the effectiveness of the pretrained backbone in accelerating optimization.

On UCF11, cross-attention starts to diverge from the other methods around epoch~5. It ultimately reaches the highest validation accuracy among all strategies. The loss curves align with this behavior: gated and weighted fusion achieve the lowest final training losses, suggesting more stable optimization dynamics, even though they do not lead in peak accuracy.

The pattern shifts slightly for UCF50. Convergence remains fast, but the curves cluster more tightly, making distinctions less pronounced. Weighted and gated fusion attain the lowest final training losses, while attention and weighted fusion deliver the strongest validation accuracy. In my view, this points to a reasonable trade-off between optimization stability and generalization performance across strategies.

\begin{table}[ht]
\centering
\caption{Top-$K$ accuracy (\%) on UCF11 and UCF50.}
\label{tab:topk}
\begin{tabular}{lcccccccc}
\toprule
& \multicolumn{4}{c}{\textbf{UCF11}} & \multicolumn{4}{c}{\textbf{UCF50}} \\
\cmidrule(lr){2-5}\cmidrule(lr){6-9}
\textbf{Method} & Top-1 & Top-2 & Top-3 & Top-5 & Top-1 & Top-2 & Top-3 & Top-5 \\
\midrule
RGB-only       & 96.2 & 97.8 & 98.1 & 98.1 & 95.8 & 97.7 & 98.4 & 99.1 \\
Flow-only      & 49.7 & 64.1 & 77.5 & 90.0 & 60.0 & 74.3 & 79.5 & 86.5 \\
\midrule
Late           & 96.2 & 98.1 & 99.4 & 99.7 & 95.2 & 97.8 & 99.0 & 99.3 \\
Concat         & 97.8 & 99.1 & 99.7 & 100.0 & 95.5 & 97.6 & 98.7 & 99.5 \\
Attention      & 98.1 & 99.4 & 99.7 & 100.0 & 96.5 & 98.8 & 99.3 & 99.5 \\
Weighted       & 96.9 & 98.1 & 99.4 & 99.7  & 96.9 & 98.3 & 99.0 & 99.6 \\
Gated          & 97.8 & 99.4 & 99.7 & 100.0 & 95.4 & 98.0 & 99.4 & 99.6 \\
\bottomrule
\end{tabular}
\end{table}

\subsection{Top-K Accuracy}

\begin{figure}[ht]
  \centering
  \includegraphics[width=\linewidth]{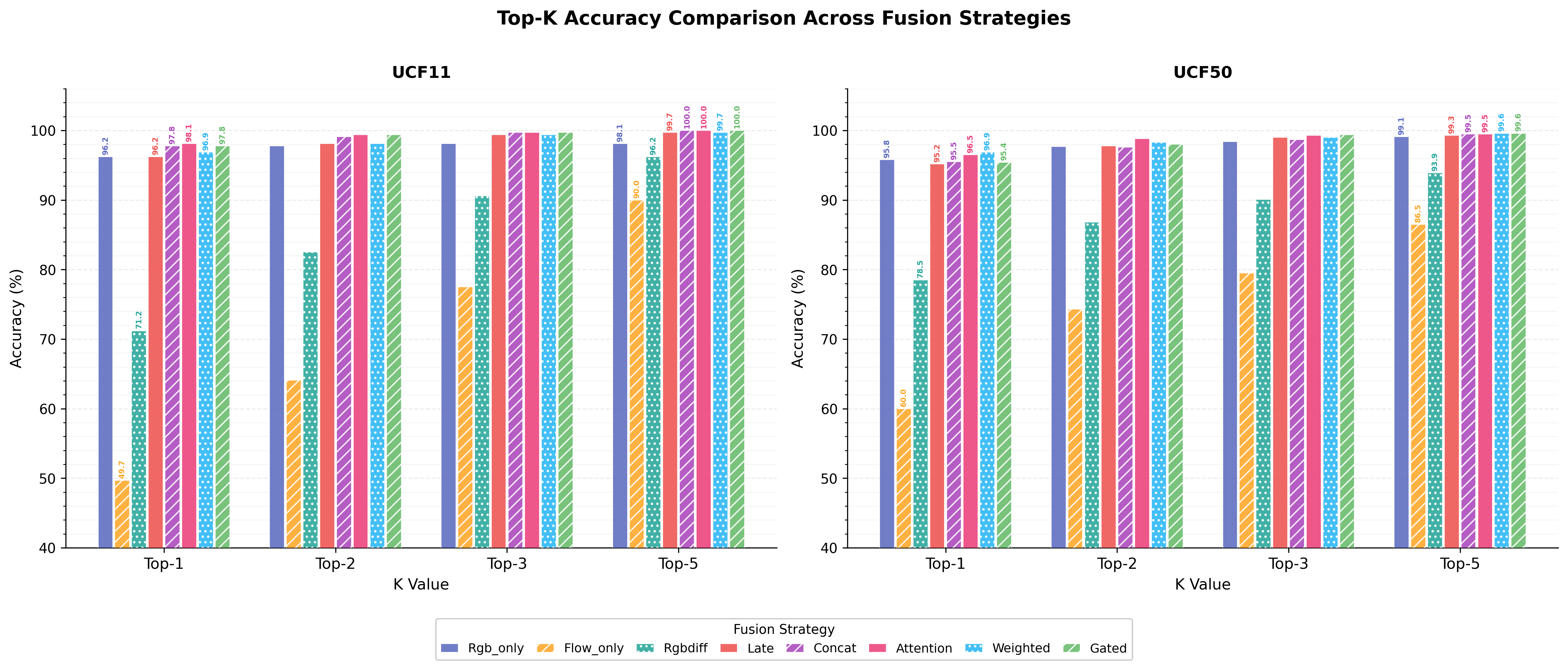}
  \caption{Top-$K$ accuracy comparison across all fusion strategies on UCF11 (left) and UCF50 (right). Hatching patterns allow distinction in greyscale. The legend
    is shared between both subplots.}
  \label{fig:topk}
\end{figure}

Table~\ref{tab:topk} and Figure~\ref{fig:topk} show Top-1 through Top-5 accuracy for all configurations on both datasets. On UCF11, most dual-stream strategies exceed 99\% at Top-3. Concatenation, cross-attention, and gated 
fusion all reach 100\% at Top-5 --- meaning the correct class always appears among the top five predictions for those models. The large gap for the flow-only model (49.7\% at Top-1, 90.0\% at Top-5) reflects how difficult it is to rank classes correctly from motion features alone when flow quality is degraded by video compression.
On UCF50, Top-$K$ accuracy rises quickly across strategies, with most exceeding 99\% by Top-3. The most telling gap is for the flow-only model: a 26.5 percentage point jump from Top-1 (60.0\%) to Top-5 (86.5\%). This suggests the model frequently has the correct class somewhere in its top predictions --- just not at the top. Motion features, under compression-degraded flow, struggle to discriminate confidently rather than to recognise entirely.

\subsection{Cross-Dataset Comparison}

Table~\ref{tab:cross} shows test accuracy for all configurations on both 
datasets side by side. The most striking pattern is the flow-only model's improvement of 10.30 percentage points from UCF11 to UCF50 --- the largest gain of any configuration. This is expected: a model trained from scratch benefits most from a larger training set, with no pretrained features to fall back on.

The RGB-only and dual-stream models, by contrast, stay roughly stable or decline slightly, reflecting the increased difficulty of the 50-class problem. Among the dual-stream strategies, weighted fusion stands out. Its test accuracy barely changes between the two datasets ($-0.02$ percentage points), while all other strategies drop by between 1.04 and 2.45 points. The reason is straightforward: weighted fusion adds only two extra parameters beyond the classifier, which are easy to optimise regardless of dataset size. Strategies with more fusion parameters --- concatenation and gated fusion --- drop the most, likely because their additional parameters remain undertrained within the 7-epoch UCF50 budget.

\begin{table}[ht]
\centering
\caption{Test accuracy on UCF11 and UCF50 for all configurations. The bottom rows show the learned stream weights from weighted fusion on each dataset.}
\label{tab:cross}
\begin{tabular}{lcc}
\toprule
\textbf{Configuration} & \textbf{UCF11 Test (\%)} & \textbf{UCF50 Test (\%)} \\
\midrule
RGB-only         & 95.94 & 95.81 \\
Flow-only        & 49.69 & 59.99 \\
\midrule
Late fusion      & 96.25 & 95.21 \\
Concat fusion    & 97.81 & 95.51 \\
Attention fusion & \textbf{98.12} & 96.48 \\
Weighted fusion  & 96.88 & \textbf{96.86} \\
Gated fusion     & 97.81 & 95.36 \\
\midrule
$\alpha_\text{RGB}$ (weighted)  & 0.507 & 0.554 \\
$\alpha_\text{flow}$ (weighted) & 0.493 & 0.446 \\
\bottomrule
\end{tabular}
\end{table}

\subsection{Confusion Matrix Analysis}

\begin{figure}[htbp]
  \centering
  \begin{subfigure}[b]{0.48\linewidth}
    \includegraphics[width=\linewidth,
    trim=0cm 0cm 28cm 1cm,
    clip]{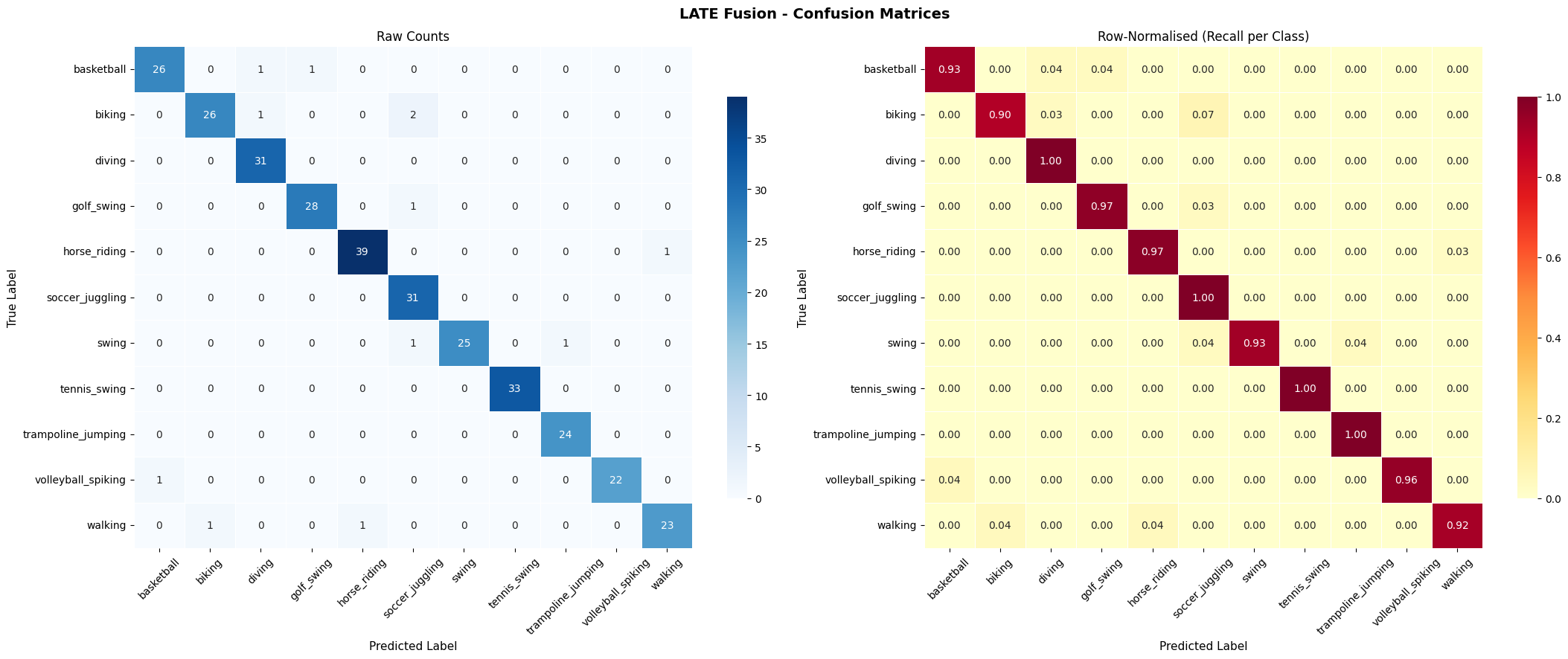}
    \caption{Late fusion (96.25\%)}
  \end{subfigure}
  \hfill
  \begin{subfigure}[b]{0.48\linewidth}
    \includegraphics[width=\linewidth,
    trim=0cm 0cm 28cm 1cm,
    clip]{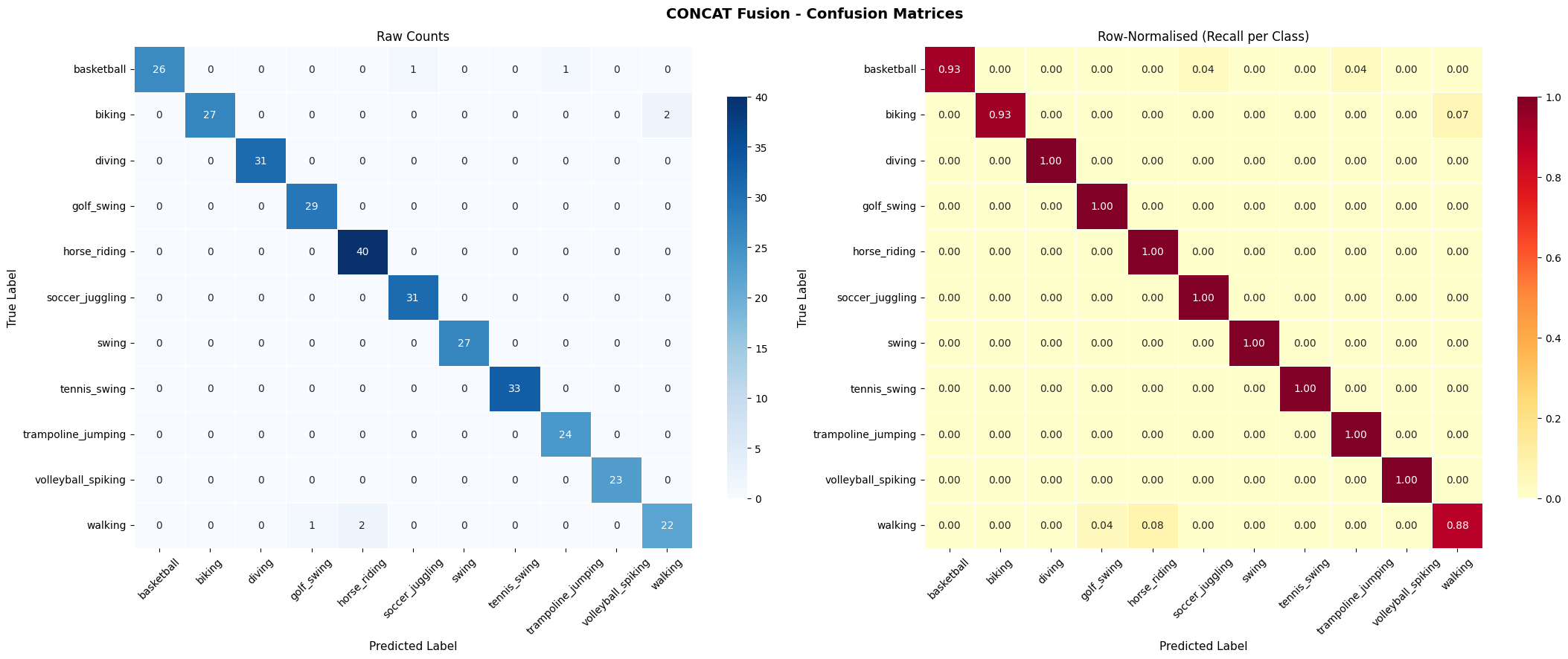}
    \caption{Concat fusion (97.81\%)}
  \end{subfigure}
  \vspace{0.5em}
  \begin{subfigure}[b]{0.48\linewidth}
    \includegraphics[width=\linewidth,
    trim=0cm 0cm 28cm 1cm,
    clip]{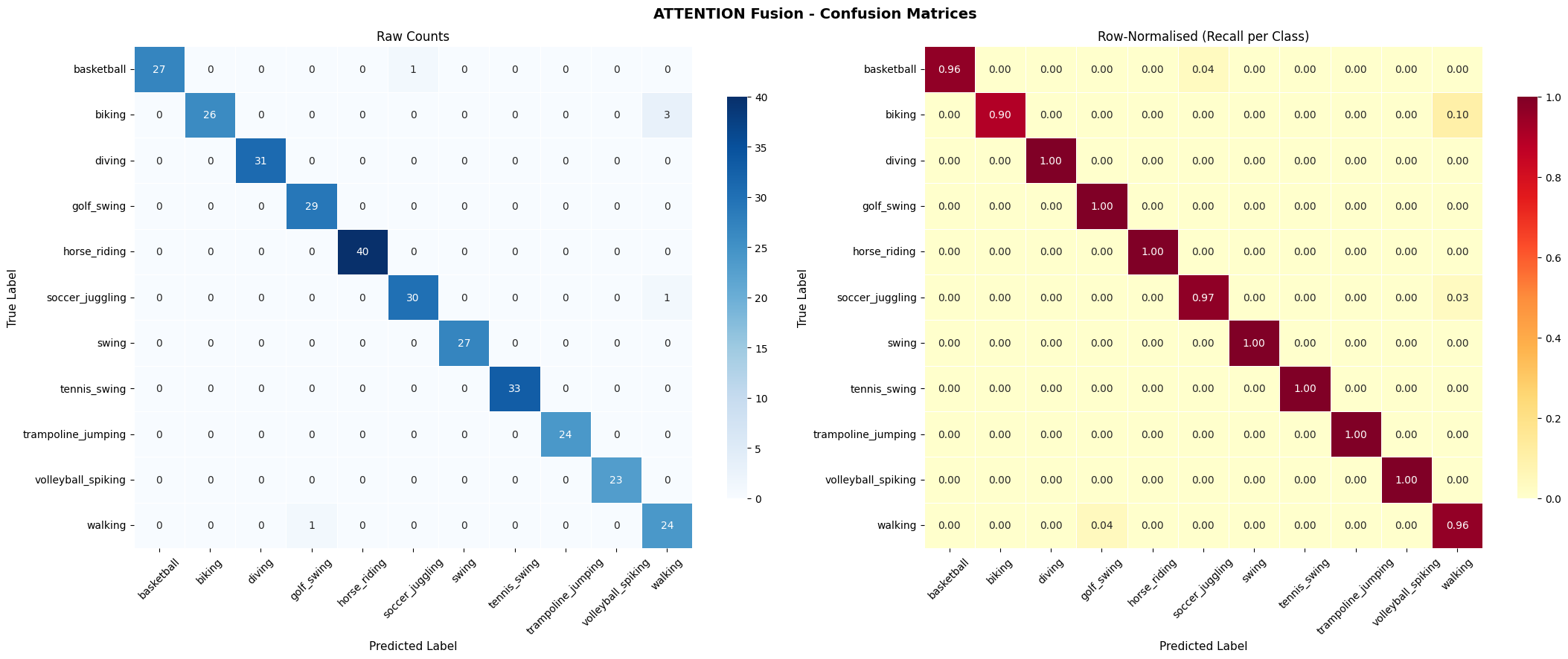}
    \caption{Attention fusion (98.12\%)}
  \end{subfigure}
  \hfill
  \begin{subfigure}[b]{0.48\linewidth}
    \includegraphics[width=\linewidth,
    trim=0cm 0cm 28cm 1cm,
    clip]{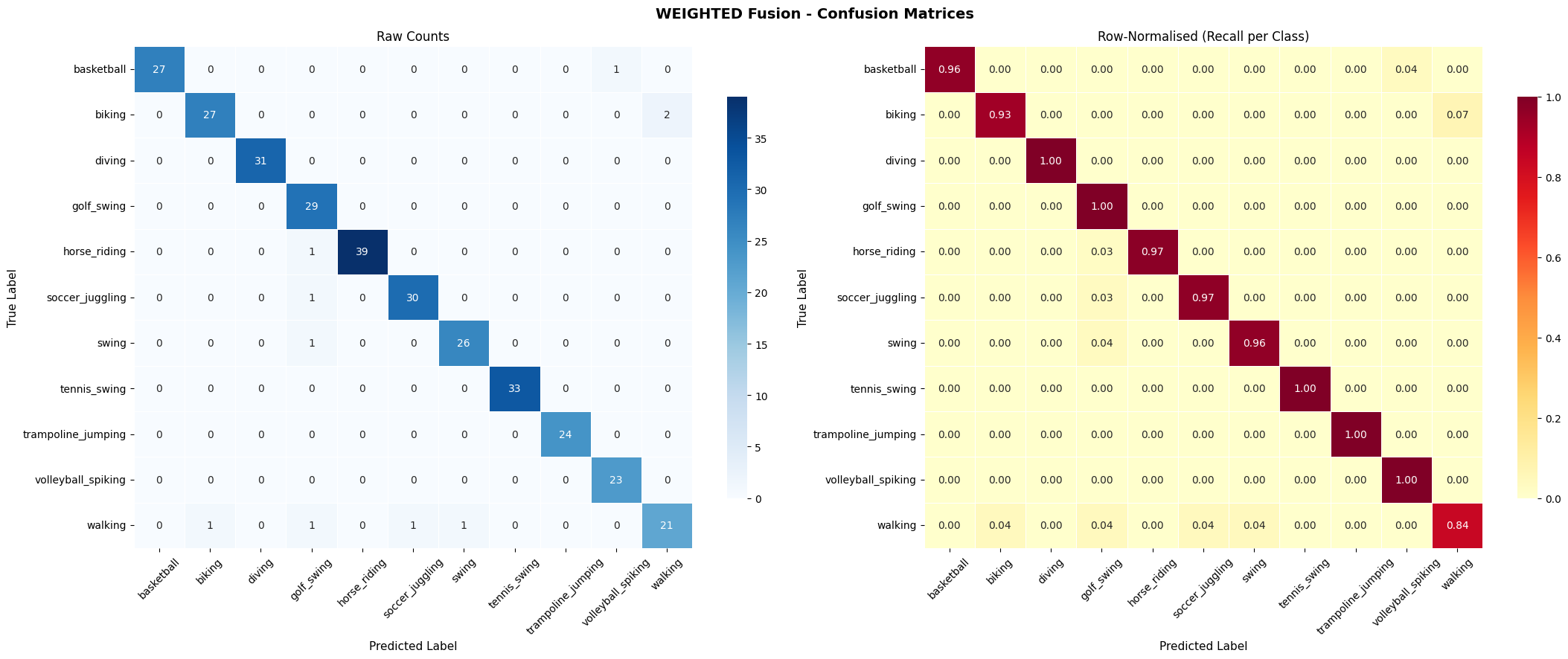}
    \caption{Weighted fusion (96.88\%)}
  \end{subfigure}
  \vspace{0.5em}
  \begin{subfigure}[b]{0.48\linewidth}
    \includegraphics[width=\linewidth,
    trim=0cm 0cm 28cm 1cm,
    clip]{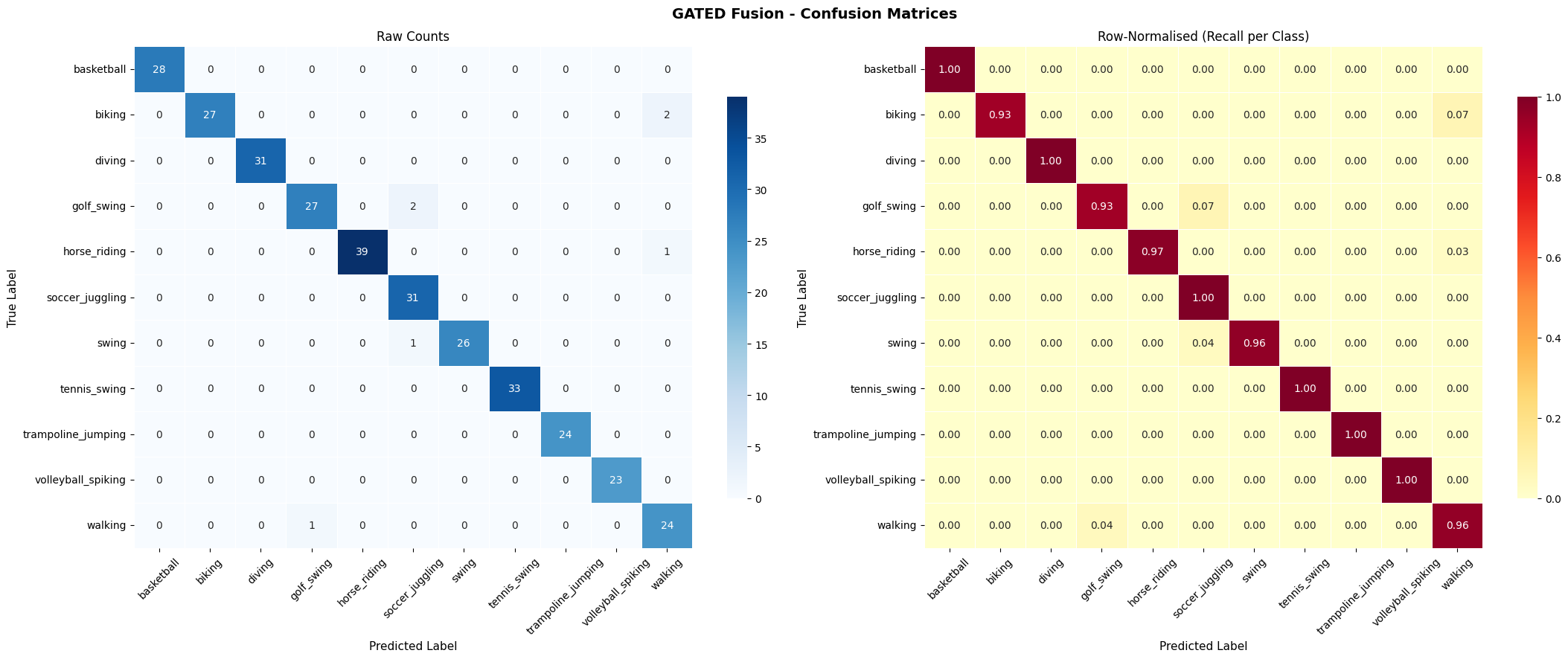}
    \caption{Gated fusion (97.81\%)}
  \end{subfigure}
  \caption{Confusion matrices (row-normalised) for all five dual-stream fusion
    strategies on UCF11. Test accuracy is shown in parentheses for each strategy.
    The strong diagonal in all cases confirms reliable per-class discrimination.
    Misclassifications are mostly limited to visually similar action pairs.}
  \label{fig:cm_ucf11}
\end{figure}

Figure~\ref{fig:cm_ucf11} shows the per-class confusion matrices for all five dual-stream fusion strategies on UCF11. In all cases, the diagonal is strongly dominant,
reflecting high overall accuracy. The small number of misclassifications tends to involve visually or motionally similar class pairs---for example, swing and tennis swing, or
horse riding and golf swing---which is understandable given the single-frame appearance input. Cross-attention has the fewest off-diagonal entries overall, consistent with its
best test accuracy of 98.12\%. Weighted fusion and late fusion show slightly more off-diagonal errors, mainly concentrated in a few similar action pairs.

\subsection{Comparison with State-of-the-Art}

Table~\ref{tab:sota} presents a comparative evaluation of DualStreamHybrid against prior methods on UCF11 and UCF50. The results demonstrate that the proposed architecture achieves strong performance on both benchmarks across multiple fusion configurations.
On UCF11, early handcrafted approaches such as Liu~\etal~\cite{liu2009recognizing}, which combined HOG/HOF descriptors with an SVM classifier, achieved 71.2\%. Subsequent deep learning methods progressively closed this gap---Yang~\etal~\cite{yang2018learning} reported 85.3\% using a Block-Term LSTM, and Gammulle~\etal~\cite{gammulle2017two} reached 94.6\% through a two-stream LSTM built on homogeneous VGG-16 backbones. Our RGB-only ViT-Tiny baseline surpasses this at 95.94\%, and the cross-attention fusion variant of DualStreamHybrid achieves 98.12\%---an improvement of 3.52 percentage points over Gammulle~\etal. The highest reported result on UCF11 belongs to Mahmoud and Abdelhakim~\cite{kumar2023attention} at 98.9\%, marginally above ours; notably, however, that method operates on RGB alone without any motion stream.

\begin{table}[ht]
\centering
\caption{Comparison with prior methods that report results on UCF11 and/or UCF50. N/A indicates the benchmark was not evaluated by that method.}
\label{tab:sota}
\renewcommand{\arraystretch}{1.15}
\begin{tabular}{llcc}
\toprule
\textbf{Method} & \textbf{Approach} & \textbf{UCF11 (\%)} & \textbf{UCF50 (\%)} \\
\midrule
Liu~\etal~\cite{liu2009recognizing} (2009)
  & HOG/HOF + SVM             & 71.2  & N/A   \\
Reddy \& Shah~\cite{reddy2013recognizing} (2013)
  & Action Bank features       & N/A   & 57.9  \\
Yang~\etal~\cite{yang2018learning} (2018)
  & Block-Term LSTM            & 85.3  & N/A   \\
Gammulle~\etal~\cite{gammulle2017two} (2017)
  & Two-Stream LSTM (VGG-16)   & 94.6  & N/A   \\
Hussain~\etal~\cite{hussain2022vision} (2022)
  & ViT-Base/16 + LSTM         & N/A   & 96.1  \\
Mahmoud \& Abdelhakim~\cite{kumar2023attention} (2023)
  & BiLSTM + Attention         & 98.9  & 96.04 \\
\midrule
RGB-only ViT-Tiny (ours)
  & ViT-Tiny/16, single stream & 95.94 & 95.81 \\
\textbf{DualStreamHybrid -- Attention (ours)}
  & ViT-Tiny + MobileNetV2, cross-attn & \textbf{98.12} & 96.48 \\
\textbf{DualStreamHybrid -- Weighted (ours)}
  & ViT-Tiny + MobileNetV2, weighted   & 96.88 & \textbf{96.86} \\
\bottomrule
\end{tabular}
\end{table}

On UCF50, the progression is similarly clear. Reddy and Shah~\cite{reddy2013recognizing} established an early baseline of 57.9\% using Action Bank features. Hussain~\etal~\cite{hussain2022vision} later reported 96.1\% with a ViT-Base/16 + LSTM architecture, and Mahmoud and Abdelhakim~\cite{kumar2023attention} achieved 96.04\% via BiLSTM with attention. Our weighted fusion variant attains 96.86\%, outperforming both prior methods while additionally incorporating a dedicated optical flow stream---a modality neither baseline employs. This indicates that the heterogeneous two-stream design of DualStreamHybrid is effective, and that further improvements in motion estimation quality could yield additional gains.
\subsection{Discussion}

The results point to several observations worth discussing. The most practically useful finding is that the best fusion strategy depends on dataset scale. Cross-attention works best on UCF11, where the attention mechanism has enough data to learn something useful within 10 epochs across 11 classes. On UCF50, the same mechanism needs more training than the 7-epoch budget allows, and weighted fusion --- simpler to train, and more parameter-efficient --- performs better and more consistently. For practitioners applying this framework to a new setting without a clear sense of data scale or training budget, weighted fusion is the safer default.

The Farneback flow bottleneck deserves separate attention. The flow-only model 
achieving only 49.69\% on UCF11 is not because optical flow is uninformative in principle --- it is because the Farneback estimator degrades under the MPEG/AVI compression artefacts present in both datasets. This points toward two practical remedies: preprocessing videos before flow extraction, or switching to a learning-based estimator like RAFT~\cite{teed2020raft} that is more robust to compression. That said, combining Farneback flow with ViT appearance in the dual-stream setup still improves over appearance alone. Motion information, even in degraded form, remains useful.

The learned stream weights in weighted fusion tell an interesting story. The 
shift from near-equal weights on UCF11 to a moderate appearance preference on 
UCF50 --- $\alpha_\text{RGB}$ rising from 0.507 to 0.554 --- is a small but 
consistent change. As the action space grows larger and more visually diverse, 
appearance features carry relatively more discriminative information, and the 
weights reflect this naturally. Crucially, $\alpha_\text{flow}$ never drops 
close to zero on either dataset, confirming that the motion stream contributes 
genuinely throughout --- even when it is not the dominant modality.

\section{Conclusion}
\label{sec:conclusion}

We proposed \emph{DualStreamHybrid}, a heterogeneous two-stream architecture 
for video action recognition that pairs a pretrained ViT-Tiny/16 for RGB 
appearance with a MobileNetV2 for optical flow, connected by a learned 
projection layer. Five fusion strategies --- late fusion, concatenation, 
cross-attention, weighted fusion, and gated fusion --- were designed and 
evaluated in a controlled ablation on UCF11 and UCF50.

The findings are clear. Cross-attention fusion achieves the best test accuracy 
on UCF11 at 98.12\%, outperforming the RGB-only ViT-Tiny baseline by roughly 2 
percentage points. On UCF50, weighted fusion reaches 96.86\% and proves 
uniquely stable across both datasets, shifting by only 0.02 percentage points 
between them. The learned stream weights confirm that both modalities contribute meaningfully on both datasets, with the RGB stream carrying slightly more weight on the larger UCF50. The flow-only stream is constrained by Farneback flow quality under compressed video conditions --- yet even degraded motion features help when combined with the RGB stream.

\textbf{Limitations.} All experiments used a single random seed, which limits 
confidence in small performance differences. A batch size of 8 likely introduces gradient noise that disproportionately affects more complex fusion mechanisms. The 7-epoch schedule on UCF50 probably leaves some strategies undertrained, and the UCF50 conclusions should be read with appropriate caution. The single-frame RGB input also discards temporal information that could meaningfully improve recognition.

\textbf{Future work.} The most pressing next step is replacing Farneback flow 
with a learning-based estimator --- RAFT~\cite{teed2020raft} or 
PWC-Net~\cite{sun2018pwc} --- to address the motion quality bottleneck directly. Evaluating on larger benchmarks such as UCF101~\cite{soomro2012ucf101} and Kinetics-400~\cite{kay2017kinetics} would test how the framework scales beyond the two datasets used here. Running experiments across multiple random seeds would make comparisons statistically more robust. Extending the RGB stream to process short clips rather than single frames would add temporal modelling to the appearance side --- an extension we consider particularly promising.

\bibliographystyle{abbrvnat}
\bibliography{references}

\begin{thebibliography}{36}
\providecommand{\natexlab}[1]{#1}
\providecommand{\url}[1]{\texttt{#1}}
\expandafter\ifx\csname urlstyle\endcsname\relax
  \providecommand{\doi}[1]{doi: #1}\else
  \providecommand{\doi}{doi: \begingroup \urlstyle{rm}\Url}\fi

\bibitem[Aggarwal and Ryoo(2011)]{aggarwal2011human}
J.~K. Aggarwal and M.~S. Ryoo.
\newblock Human activity analysis: A review.
\newblock \emph{ACM Computing Surveys}, 43\penalty0 (3):\penalty0 16:1--16:43,
  2011.
\newblock \doi{10.1145/1922649.1922653}.

\bibitem[Alamri et~al.(2019)Alamri, Cartillier, Das, Wang, Cherian, Essa,
  Batra, Marks, Hori, Anderson, Lee, and Parikh]{Alamri_2019_CVPR}
H.~Alamri, V.~Cartillier, A.~Das, J.~Wang, A.~Cherian, I.~Essa, D.~Batra, T.~K.
  Marks, C.~Hori, P.~Anderson, S.~Lee, and D.~Parikh.
\newblock Audio visual scene-aware dialog.
\newblock In \emph{Proceedings of the IEEE/CVF Conference on Computer Vision
  and Pattern Recognition (CVPR)}, June 2019.
\newblock \doi{10.1109/CVPR.2019.00774}.

\bibitem[Arnab et~al.(2021)Arnab, Dehghani, Heigold, Sun, Lu{\v{c}}i{\'{c}},
  and Schmid]{arnab2021vivit}
A.~Arnab, M.~Dehghani, G.~Heigold, C.~Sun, M.~Lu{\v{c}}i{\'{c}}, and C.~Schmid.
\newblock {ViViT}: {A} video vision transformer.
\newblock In \emph{IEEE/CVF International Conference on Computer Vision
  (ICCV)}, pages 6836--6846, 2021.
\newblock \doi{10.1109/ICCV48922.2021.00676}.

\bibitem[Baladaniya and Choudhary(2025)]{baladaniya2025artificial}
M.~Baladaniya and A.~K. Choudhary.
\newblock Artificial intelligence in sports science: A systematic review on
  performance optimization, injury prevention, and rehabilitation.
\newblock \emph{Journal of Clinical Medicine of Kazakhstan}, 22\penalty0
  (3):\penalty0 64--72, 2025.
\newblock \doi{10.23950/jcmk/16412}.

\bibitem[Bertasius et~al.(2021)Bertasius, Wang, and
  Torresani]{bertasius2021space}
G.~Bertasius, H.~Wang, and L.~Torresani.
\newblock Is space-time attention all you need for video understanding?
\newblock In \emph{International Conference on Machine Learning (ICML)}, volume
  139, pages 813--824, 2021.

\bibitem[Carreira and Zisserman(2017)]{carreira2017quo}
J.~Carreira and A.~Zisserman.
\newblock Quo vadis, action recognition? {A} new model and the {Kinetics}
  dataset.
\newblock In \emph{IEEE Conference on Computer Vision and Pattern Recognition
  (CVPR)}, pages 6299--6308, 2017.
\newblock \doi{10.1109/CVPR.2017.502}.

\bibitem[Deng et~al.(2009)Deng, Dong, Socher, Li, Li, and
  Fei-Fei]{deng2009imagenet}
J.~Deng, W.~Dong, R.~Socher, L.-J. Li, K.~Li, and L.~Fei-Fei.
\newblock {ImageNet}: A large-scale hierarchical image database.
\newblock In \emph{IEEE Conference on Computer Vision and Pattern Recognition
  (CVPR)}, pages 248--255, 2009.
\newblock \doi{10.1109/CVPR.2009.5206848}.

\bibitem[Dosovitskiy et~al.(2021)Dosovitskiy, Beyer, Kolesnikov, Weissenborn,
  Zhai, Unterthiner, Dehghani, Minderer, Heigold, Gelly, Uszkoreit, and
  Houlsby]{dosovitskiy2020image}
A.~Dosovitskiy, L.~Beyer, A.~Kolesnikov, D.~Weissenborn, X.~Zhai,
  T.~Unterthiner, M.~Dehghani, M.~Minderer, G.~Heigold, S.~Gelly, J.~Uszkoreit,
  and N.~Houlsby.
\newblock An image is worth 16x16 words: Transformers for image recognition at
  scale.
\newblock In \emph{International Conference on Learning Representations
  (ICLR)}, 2021.

\bibitem[Fan et~al.(2021)Fan, Xiong, Mangalam, Li, Yan, Malik, and
  Feichtenhofer]{fan2021multiscale}
H.~Fan, B.~Xiong, K.~Mangalam, Y.~Li, Z.~Yan, J.~Malik, and C.~Feichtenhofer.
\newblock Multiscale vision transformers.
\newblock In \emph{IEEE/CVF International Conference on Computer Vision
  (ICCV)}, pages 6824--6835, 2021.
\newblock \doi{10.1109/ICCV48922.2021.00675}.

\bibitem[Farneb{\"{a}}ck(2003)]{farneback2003two}
G.~Farneb{\"{a}}ck.
\newblock Two-frame motion estimation based on polynomial expansion.
\newblock In \emph{Scandinavian Conference on Image Analysis (SCIA)}, pages
  363--370, 2003.
\newblock \doi{10.1007/3-540-45103-X_50}.

\bibitem[Feichtenhofer et~al.(2016)Feichtenhofer, Pinz, and
  Zisserman]{feichtenhofer2016convolutional}
C.~Feichtenhofer, A.~Pinz, and A.~Zisserman.
\newblock Convolutional two-stream network fusion for video action recognition.
\newblock In \emph{IEEE Conference on Computer Vision and Pattern Recognition
  (CVPR)}, pages 1933--1941, 2016.
\newblock \doi{10.1109/CVPR.2016.213}.

\bibitem[Feichtenhofer et~al.(2019)Feichtenhofer, Fan, Malik, and
  He]{feichtenhofer2019slowfast}
C.~Feichtenhofer, H.~Fan, J.~Malik, and K.~He.
\newblock {SlowFast} networks for video recognition.
\newblock In \emph{IEEE/CVF International Conference on Computer Vision
  (ICCV)}, pages 6202--6211, 2019.
\newblock \doi{10.1109/ICCV.2019.00630}.

\bibitem[Gadzicki et~al.(2020)Gadzicki, Knappe, and
  Zetzsche]{gadzicki2020early}
K.~Gadzicki, R.~Knappe, and C.~Zetzsche.
\newblock Early vs late fusion in multimodal convolutional neural networks.
\newblock In \emph{IEEE International Conference on Information Fusion
  (FUSION)}, pages 1--7, 2020.
\newblock \doi{10.23919/FUSION45008.2020.9190246}.

\bibitem[Gammulle et~al.(2017)Gammulle, Denman, Sridharan, and
  Fookes]{gammulle2017two}
H.~Gammulle, S.~Denman, S.~Sridharan, and C.~Fookes.
\newblock Two stream {LSTM}: A deep fusion framework for human action
  recognition.
\newblock In \emph{IEEE Winter Conference on Applications of Computer Vision
  (WACV)}, pages 177--186, 2017.
\newblock \doi{10.1109/WACV.2017.27}.

\bibitem[He et~al.(2016)He, Zhang, Ren, and Sun]{he2016deep}
K.~He, X.~Zhang, S.~Ren, and J.~Sun.
\newblock Deep residual learning for image recognition.
\newblock In \emph{IEEE Conference on Computer Vision and Pattern Recognition
  (CVPR)}, pages 770--778, 2016.
\newblock \doi{10.1109/CVPR.2016.90}.

\bibitem[Hussain et~al.(2022)Hussain, Muhammad, Ullah, Ahmad, Baik, and
  de~Albuquerque]{hussain2022vision}
A.~Hussain, K.~Muhammad, A.~Ullah, Z.~Ahmad, S.~W. Baik, and V.~H.~C.
  de~Albuquerque.
\newblock Vision transformer and deep sequence learning for human activity
  recognition in surveillance videos.
\newblock \emph{Computational Intelligence and Neuroscience}, 2022:\penalty0
  3454167, 2022.
\newblock \doi{10.1155/2022/3454167}.

\bibitem[Ilse et~al.(2018)Ilse, Tomczak, and Welling]{ilse2018attention}
M.~Ilse, J.~M. Tomczak, and M.~Welling.
\newblock Attention-based deep multiple instance learning.
\newblock In \emph{International Conference on Machine Learning (ICML)},
  volume~80, pages 2132--2141, 2018.

\bibitem[Karpathy et~al.(2014)Karpathy, Toderici, Shetty, Leung, Sukthankar,
  and Fei-Fei]{karpathy2014large}
A.~Karpathy, G.~Toderici, S.~Shetty, T.~Leung, R.~Sukthankar, and L.~Fei-Fei.
\newblock Large-scale video classification with convolutional neural networks.
\newblock In \emph{IEEE Conference on Computer Vision and Pattern Recognition
  (CVPR)}, pages 1725--1732, 2014.
\newblock \doi{10.1109/CVPR.2014.223}.

\bibitem[Kay et~al.(2017)Kay, Carreira, Simonyan, Zhang, Hillier,
  Vijayanarasimhan, Viola, Green, Back, Natsev, Suleyman, and
  Zisserman]{kay2017kinetics}
W.~Kay, J.~Carreira, K.~Simonyan, B.~Zhang, C.~Hillier, S.~Vijayanarasimhan,
  F.~Viola, T.~Green, T.~Back, P.~Natsev, M.~Suleyman, and A.~Zisserman.
\newblock The {Kinetics} human action video dataset.
\newblock \emph{arXiv preprint arXiv:1705.06950}, 2017.

\bibitem[Kumar et~al.(2023)Kumar, Patel, Biswas, and
  Shitharth]{kumar2023attention}
M.~Kumar, A.~K. Patel, M.~Biswas, and S.~Shitharth.
\newblock Attention-based bidirectional-long short-term memory for abnormal
  human activity detection.
\newblock \emph{Scientific Reports}, 13\penalty0 (1):\penalty0 14442, 2023.
\newblock \doi{10.1038/s41598-023-41231-0}.

\bibitem[Liu et~al.(2009)Liu, Luo, and Shah]{liu2009recognizing}
J.~Liu, J.~Luo, and M.~Shah.
\newblock Recognizing realistic actions from videos ``in the wild''.
\newblock In \emph{IEEE Conference on Computer Vision and Pattern Recognition
  (CVPR)}, pages 1996--2003, 2009.
\newblock \doi{10.1109/CVPR.2009.5206744}.

\bibitem[Liu et~al.(2021)Liu, Lin, Cao, Hu, Wei, Zhang, Lin, and
  Guo]{liu2021swin}
Z.~Liu, Y.~Lin, Y.~Cao, H.~Hu, Y.~Wei, Z.~Zhang, S.~Lin, and B.~Guo.
\newblock Swin transformer: Hierarchical vision transformer using shifted
  windows.
\newblock In \emph{IEEE/CVF International Conference on Computer Vision
  (ICCV)}, pages 10012--10022, 2021.
\newblock \doi{10.1109/ICCV48922.2021.00986}.

\bibitem[Liu et~al.(2022)Liu, Ning, Cao, Wei, Zhang, Lin, and Hu]{liu2022video}
Z.~Liu, J.~Ning, Y.~Cao, Y.~Wei, Z.~Zhang, S.~Lin, and H.~Hu.
\newblock Video {Swin} transformer.
\newblock In \emph{IEEE/CVF Conference on Computer Vision and Pattern
  Recognition (CVPR)}, pages 3202--3211, 2022.
\newblock \doi{10.1109/CVPR52688.2022.00320}.

\bibitem[Loshchilov and Hutter(2019)]{loshchilov2017decoupled}
I.~Loshchilov and F.~Hutter.
\newblock Decoupled weight decay regularization.
\newblock In \emph{International Conference on Learning Representations
  (ICLR)}, 2019.

\bibitem[Poppe(2010)]{poppe2010survey}
R.~Poppe.
\newblock A survey on vision-based human action recognition.
\newblock \emph{Image and Vision Computing}, 28\penalty0 (6):\penalty0
  976--990, 2010.
\newblock \doi{10.1016/j.imavis.2009.11.014}.

\bibitem[Reddy and Shah(2013)]{reddy2013recognizing}
K.~K. Reddy and M.~Shah.
\newblock Recognizing 50 human action categories of web videos.
\newblock \emph{Machine Vision and Applications}, 24\penalty0 (5):\penalty0
  971--981, 2013.
\newblock \doi{10.1007/s00138-012-0450-4}.

\bibitem[Sandler et~al.(2018)Sandler, Howard, Zhu, Zhmoginov, and
  Chen]{sandler2018mobilenetv2}
M.~Sandler, A.~Howard, M.~Zhu, A.~Zhmoginov, and L.-C. Chen.
\newblock {MobileNetV2}: Inverted residuals and linear bottlenecks.
\newblock In \emph{IEEE Conference on Computer Vision and Pattern Recognition
  (CVPR)}, pages 4510--4520, 2018.
\newblock \doi{10.1109/CVPR.2018.00474}.

\bibitem[Simonyan and Zisserman(2014)]{simonyan2014two}
K.~Simonyan and A.~Zisserman.
\newblock Two-stream convolutional networks for action recognition in videos.
\newblock In \emph{Advances in Neural Information Processing Systems
  (NeurIPS)}, volume~27, pages 568--576, 2014.
\newblock \doi{10.48550/arXiv.1406.2199}.

\bibitem[Simonyan and Zisserman(2015)]{simonyan2014vgg}
K.~Simonyan and A.~Zisserman.
\newblock Very deep convolutional networks for large-scale image recognition.
\newblock In \emph{International Conference on Learning Representations
  (ICLR)}, 2015.
\newblock \doi{10.48550/arXiv.1409.1556}.

\bibitem[Soomro et~al.(2012)Soomro, Zamir, and Shah]{soomro2012ucf101}
K.~Soomro, A.~R. Zamir, and M.~Shah.
\newblock {UCF101}: A dataset of 101 human actions classes from videos in the
  wild.
\newblock Technical report, CRCV Technical Report CRCV-TR-12-01, 2012.

\bibitem[Sun et~al.(2018)Sun, Yang, Liu, and Kautz]{sun2018pwc}
D.~Sun, X.~Yang, M.-Y. Liu, and J.~Kautz.
\newblock {PWC-Net}: {CNNs} for optical flow using pyramid, warping, and cost
  volume.
\newblock In \emph{IEEE Conference on Computer Vision and Pattern Recognition
  (CVPR)}, pages 8934--8943, 2018.
\newblock \doi{10.1109/CVPR.2018.00931}.

\bibitem[Teed and Deng(2020)]{teed2020raft}
Z.~Teed and J.~Deng.
\newblock {RAFT}: Recurrent all-pairs field transforms for optical flow.
\newblock In \emph{European Conference on Computer Vision (ECCV)}, pages
  402--419, 2020.
\newblock \doi{10.1007/978-3-030-58536-5_24}.

\bibitem[Tran et~al.(2015)Tran, Bourdev, Fergus, Torresani, and
  Paluri]{tran2015learning}
D.~Tran, L.~Bourdev, R.~Fergus, L.~Torresani, and M.~Paluri.
\newblock Learning spatiotemporal features with {3D} convolutional networks.
\newblock In \emph{IEEE International Conference on Computer Vision (ICCV)},
  pages 4489--4497, 2015.
\newblock \doi{10.1109/ICCV.2015.510}.

\bibitem[Wang et~al.(2016)Wang, Xiong, Wang, Qiao, Lin, Tang, and {Van
  Gool}]{wang2016temporal}
L.~Wang, Y.~Xiong, Z.~Wang, Y.~Qiao, D.~Lin, X.~Tang, and L.~{Van Gool}.
\newblock Temporal segment networks: Towards good practices for deep action
  recognition.
\newblock In \emph{European Conference on Computer Vision (ECCV)}, pages
  20--36, 2016.
\newblock \doi{10.1007/978-3-319-46484-8_2}.

\bibitem[Ye et~al.(2018)Ye, Wang, Li, Chen, Zhe, Chu, and Xu]{yang2018learning}
J.~Ye, L.~Wang, G.~Li, D.~Chen, S.~Zhe, X.~Chu, and Z.~Xu.
\newblock Learning compact recurrent neural networks with block-term tensor
  decomposition.
\newblock In \emph{IEEE Conference on Computer Vision and Pattern Recognition
  (CVPR)}, pages 9378--9387, 2018.
\newblock \doi{10.1109/CVPR.2018.00977}.

\bibitem[Zach et~al.(2007)Zach, Pock, and Bischof]{zach2007tvl1}
C.~Zach, T.~Pock, and H.~Bischof.
\newblock A duality based approach for realtime {TV-L1} optical flow.
\newblock In \emph{DAGM Symposium on Pattern Recognition}, pages 214--223,
  2007.
\newblock \doi{10.1007/978-3-540-74936-3_22}.

\end{thebibliography}

\end{document}